\documentclass[letterpaper, 10 pt, conference]{ieeeconf}  

\IEEEoverridecommandlockouts                              
\usepackage{float}
\usepackage{soul}

\usepackage{mathtools}
\usepackage{amssymb}
\usepackage{bbm}
\usepackage{graphicx}
\usepackage{multicol}
\usepackage{multirow, makecell}
\usepackage{balance}

\usepackage[export]{adjustbox}
\usepackage[bookmarks=true]{hyperref}
\usepackage[table]{xcolor}    
\hypersetup{
    colorlinks=true,
    linkcolor=blue,
    filecolor=magenta,      
    urlcolor=cyan,
}
\usepackage[raggedright]{sidecap}
\usepackage[font=small,skip=5pt]{caption}
\usepackage{subcaption}
\let\labelindent\relax
\usepackage{enumitem,kantlipsum}
\usepackage[tight-spacing=true]{siunitx}

\usepackage{tablefootnote}
\usepackage{booktabs}
\usepackage{dsfont}
\usepackage{algcompatible}
\usepackage{gensymb}
\usepackage[hang,flushmargin]{footmisc}
\usepackage{lipsum}

\makeatletter
\let\NAT@parse\undefined
\makeatother
\usepackage[numbers]{natbib}

\DeclarePairedDelimiterX{\infdivx}[2]{\big(}{\big)}{%
  #1\;\delimsize\|\;#2%
}
\definecolor{nice-red}{HTML}{E41A1C}
\definecolor{nice-orange}{HTML}{FF7F00}
\definecolor{nice-yellow}{HTML}{FFC020}
\definecolor{nice-green}{HTML}{4DAF4A}
\definecolor{nice-blue}{HTML}{377EB8}
\definecolor{nice-nice-red}{HTML}{984EA3}
\definecolor{nice-grey}{HTML}{91A3B0}
\definecolor{nice-amber}{HTML}{DAA520}
\usepackage{algorithm}
\newlength\myindent
\newcolumntype{P}[1]{>{\centering\arraybackslash}p{#1}}

\newcommand{\algorithmfootnote}[2][\footnotesize]{%
  \let\old@algocf@finish\@algocf@finish
  \def\@algocf@finish{\old@algocf@finish
    \leavevmode\rlap{\begin{minipage}{\linewidth}
    #1#2
    \end{minipage}}%
  }%
}
\allowdisplaybreaks
\usepackage{color,soul}

\usepackage{color}

\title{\LARGE \bf iGibson 1.0: A Simulation Environment for Interactive Tasks\\in Large Realistic Scenes
\vspace{-8mm}}
\author{\normalsize Bokui Shen$^{*}$, Fei Xia$^{*}$, Chengshu Li$^{*}$, Roberto Mart\'in-Mart\'in$^{*}$, Linxi Fan, Guanzhi Wang, Claudia P\'erez-D'Arpino, \\
Shyamal Buch, Sanjana Srivastava, Lyne Tchapmi, Micael Tchapmi, Kent Vainio, Josiah Wong, Li Fei-Fei, Silvio Savarese
\thanks{$^{*}$Equal contribution. All authors are with the Stanford Vision \& Learning Laboratory, Stanford University.
}

}
\begin{document}

\maketitle

\begin{abstract}
We present \textit{iGibson 1.0}, a novel simulation environment to develop robotic solutions for interactive tasks in large-scale realistic scenes. Our environment contains 15 \textit{fully interactive} home-sized scenes with 108 rooms populated with rigid and articulated objects. The scenes are replicas of real-world homes, with distribution and the layout of objects aligned to those of the real world. iGibson 1.0 integrates several key features to facilitate the study of interactive tasks: i) generation of high-quality virtual sensor signals (RGB, depth, segmentation, LiDAR, flow and so on), ii) domain randomization to change the materials of the objects (both visual and physical) and/or their shapes, iii) integrated sampling-based motion planners to generate collision-free trajectories for robot bases and arms, and iv) intuitive human-iGibson interface that enables efficient collection of human demonstrations. Through experiments, we show that the full interactivity of the scenes enables agents to learn useful visual representations that accelerate the training of downstream manipulation tasks. We also show that iGibson features enable the generalization of navigation agents, and that the human-iGibson interface and integrated motion planners facilitate efficient imitation learning of human demonstrated (mobile) manipulation behaviors. iGibson 1.0 is open-source, equipped with comprehensive examples and documentation. For more information, visit our project website: {\footnotesize\href{http://svl.stanford.edu/igibson/}{http://svl.stanford.edu/igibson/}}.
\end{abstract}

\section{Introduction}
\label{s_intro}

Simulation environments have proliferated over the last few years as a way to train robots and interactive agents in a rapid and safe manner. In these environments, agents learn to engage in physical interactions~\cite{lillicrap2015continuous,levine2016end}, navigate based on sensor signals~\cite{mirowski2016learning,parisotto2017neural,zhu2017target,shen2019situational,chen2019behavioral}, or plan long-horizon tasks~\cite{garrett2018ffrob,xu2018neural,xia2020relmogen,li2019hrl4in}. 
In simulation, agents learn to perform interactions that actively change the input sensor signals and the state of the environment towards a desired configuration, capabilities at the core of what an embodied agent needs to achieve.

However, existing simulation environments that combine physics simulation and robotic tasks often cater to a narrow set of tasks and include only clean, small-scale scenes~\cite{james2020rlbench,yu2020meta,urakami2019doorgym,xiang2020sapien,lee2019ikea,zhu2020robosuite}. The few simulation environments that include large scenes such as homes or offices either disable the possibility of interacting with the scene, focusing only on navigation (e.g. Habitat~\cite{habitat19arxiv}), or use simplified modes of interaction (e.g. AI2Thor~\cite{kolve2017ai2}, VirtualHome~\cite{puig2018virtualhome}). These simulators do not support the development of end-to-end sensorimotor control loops for tasks that require rich, continuous interaction with the scene. Such tasks are difficult to accomplish in the aforementioned simulators, and the simplified modes of interaction lead to difficulties in transferring the learned policy onto real robots.

\begin{figure}[t]
    \centering
    \includegraphics[width=0.98\columnwidth]{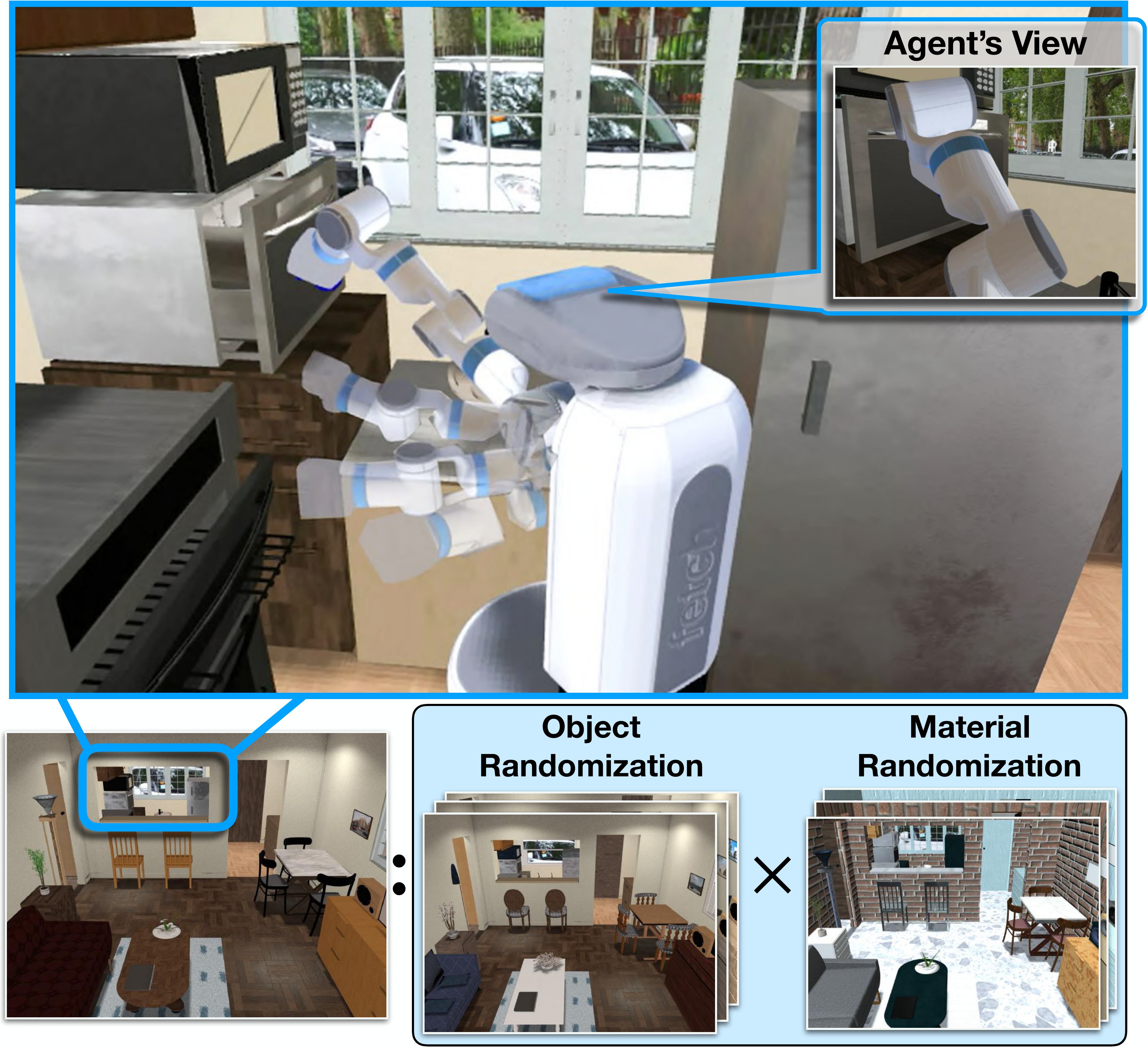}
    \vspace{-1mm}
    \caption{
    Robot performs an interactive task in iGibson 1.0. It operates in the kitchen of one of iGibson's fully interactive scenes, planning an interaction with the arm using a integrated sampling-based motion planner and receiving first-person view. \textit{Bottom:} The same scene can be randomized with different materials and/or object models
    }
    \label{fig:pullfig}
\vspace{-7mm}
\end{figure}

We present \textit{iGibson 1.0} (alternative called just \textit{iGibson} in this manuscript), a novel simulation environment that enables the development of embodied agents for interactive tasks in large-scale realistic scenes (Fig.~\ref{fig:pullfig}). Interactivity is achieved by leveraging a physics engine processing all elements in the scene, enabling manipulation of rigid and articulated objects as well as mobility.
iGibson 1.0 aims at unifying several aspects of robot simulation that are often available in different software tools, such as physics simulation for interaction with objects and robot control, high-quality simulated sensors, integration with reinforcement learning frameworks, and realistic indoor scenes that reflect the objects distribution of real homes.  This integration allows fully physics based simulation of robot tasks (i.e. simulating the full complexity of the task) and allows developing task and motion planning, reinforcement learning or imitation learning solutions for those tasks with virtual sensor signals.

iGibson 1.0 contains 15 fully interactive and visually-realistic scenes with a total of 108 rooms. These scenes were generated by annotating 3D reconstructions of real-world scans (static scenes represented by a single mesh) and converting them into fully interactive scene models (scenes filled with articulated object models).
In this process, we respect the original object-instance layout and object-category distribution. The object models are extended from open-source datasets~\cite{chang2015shapenet,wang2019shape2motion,xiang2020sapien} enriched with annotations of material and dynamic properties. iGibson's physics-based renderer leverages the extra information provided in the material annotation (maps of metallic, roughness and normals) to generate high-quality virtual images. To further facilitate the training of more robust visuomotor agents, iGibson 1.0 offers domain randomization procedures for materials (both visual appearances and dynamics properties) and object shapes while respecting the distribution of object placements and preserving interactability. iGibson 1.0 is also equipped with a graphical user interface that allows human users to easily interact with the scenes, enabling efficient collection of human demonstrations for imitation learning.

In summary, iGibson 1.0 provides the following novel features that facilitate developing and training robotic solutions:
\begin{enumerate}[wide, labelwidth=!, labelindent=0pt]

\item {Fifteen fully interactive scenes containing 108 rooms} modelled after real world homes with articulated object models annotated with materials and dynamics properties. Additionally, we support importing CubiCasa5K~\cite{kalervo2019cubicasa5k} and 3D-Front~\cite{fu20203d} layouts, giving access to more than 12000 additional fully interactive scenes.
\item Realistic virtual sensor signals, 
including a physics-based renderer (PBR) for RGB images, rendering of normals, depth, point clouds, virtual LiDAR signals, and optical/scene flow.
We further integrate domain randomization functionality (visual texture, dynamics properties and object instances) that facilitates generalization to unseen scenes.
\item Useful tooling for developing robotics solutions in simulation, such as a human-computer interface for humans to provide interactive demonstrations, and sampling-based motion planners for navigation and manipulation.

\end{enumerate}

We demonstrate the benefits of these novel features in a comprehensive set of experiments in which visual agents are trained for navigation and interactive tasks. Our experiments show that iGibson 1.0 enables researchers to 1) train more robust and generalizable sensorimotor policies thanks to its realistic virtual sensor signals (including LiDAR) and domain randomization mechanisms, 2) collect human demonstrations and train imitation learning policies for manipulation and mobile manipulation tasks, and 3) learn intermediate visual representations linked to interactability of the scene that accelerate training of downstream manipulation tasks. iGibson 1.0 is open-source and academically developed,
and available on our website {\small\href{http://svl.stanford.edu/igibson/}{http://svl.stanford.edu/igibson/}}.
\section{Related Work}
\label{s_rw}

\begin{table*}[th!]
\centering
\caption{Comparison of Simulation Environments}
\label{t:simenv}
\rowcolors{2}{gray!25}{white}
  \begin{tabular}{l|c|cccccc} 
  \toprule
     &  iGibson 1.0 (ours) & Gibson~\cite{xia2018gibson}& Habitat~\cite{habitat19arxiv} & Sapien~\cite{xiang2020sapien} & AI2Thor~\cite{kolve2017ai2} & VirtualH~\cite{puig2018virtualhome} & TDW~\cite{gan2020threedworld}  \\ 
  \midrule
  \begin{tabular}{@{}c@{}}\textbf{Provided Large Scenes} \\
  Real-World / Designed\end{tabular}
 & \begin{tabular}{@{}c@{}} 15 homes \\(108 rooms)   /  -- \end{tabular} 
 & 1400 / -- & -- & -- & -- / 120 rooms & -- / 7 & -- / 25  \\
 \begin{tabular}{@{}c@{}}\textbf{Provided Objects} \\ 
 Number / Materials\end{tabular} 
 & 570 / Yes & -- / -- & -- / -- & 2346 / No & 609 / Yes & 308 / No & 200 / Yes  \\
\begin{tabular}{@{}c@{}}\textbf{Agent/World Interaction} \\ \textbf{F}orces, \textbf{P}redefined \textbf{A}ctions \end{tabular}
 & F & -- & -- & F & F \& PA & F \& PA & F \\
\begin{tabular}{@{}c@{}}\textbf{Physics Engine}\end{tabular}
& Bullet & Bullet & Bullet & PhysX & Unity & Unity & Unity \& Flex  \\
\begin{tabular}{@{}c@{}}\textbf{Type of Simulation} \\
\textbf{R}igid \textbf{B}odies \textbf{P}hysics, \\ \textbf{R}igid \textbf{B}odies with \textbf{PA}, \\
\textbf{P}articles and \textbf{F}luids
\end{tabular}
& \begin{tabular}{@{}c@{}}RBP \end{tabular} & \begin{tabular}{@{}c@{}}RBP \end{tabular} 
& \begin{tabular}{@{}c@{}}RBP \end{tabular}
& \begin{tabular}{@{}c@{}}RBP \end{tabular}
&
\begin{tabular}{@{}c@{}}
\begin{tabular}{@{}c@{}}RBP\&RBPA  \end{tabular}
 \end{tabular}  
& \begin{tabular}{@{}c@{}} RBPA  \end{tabular} 
& \begin{tabular}{@{}c@{}}RBP\&PF\end{tabular}  \\
\begin{tabular}{@{}c@{}}\textbf{Supported Task}
\end{tabular}
& \begin{tabular}{@{}c@{}}Nav.\&Manip.\end{tabular} & \begin{tabular}{@{}c@{}}Nav.\end{tabular} 
& \begin{tabular}{@{}c@{}}Nav.\end{tabular}
& \begin{tabular}{@{}c@{}}Nav.\&Manip.\end{tabular}
&
\begin{tabular}{@{}c@{}}
\begin{tabular}{@{}c@{}}Nav.\&Manip. \end{tabular}
 \end{tabular}  
& \begin{tabular}{@{}c@{}} Nav.\&Manip. \end{tabular} 
& \begin{tabular}{@{}c@{}} Nav.\&Manip.\end{tabular}  \\
\begin{tabular}{@{}c@{}}\textbf{Type of Rendering}  \\ \end{tabular}
& PBR& IBR& PBR & PBR,RTX & PBR & PBR & PBR  \\
\textbf{Virtual Sensor Signals} & \begin{tabular}{@{}c@{}}RGB,D,N\\SS,FL,LiDAR\end{tabular}& \begin{tabular}{@{}c@{}}RGB,D\\N,SS\end{tabular}& RGB,D,SS,S  & RGB,D,SS & RGB,D,SS,S & \begin{tabular}{@{}c@{}}RGB,D\\SS,FL\end{tabular} & RGB,D,SS,S  \\
\begin{tabular}{@{}c@{}}\textbf{Domain Randomization} \\
\textbf{S}cene,\textbf{O}bject,\textbf{M}aterials\end{tabular}
 & S,O,M& - & - & - & S & S & S,O  \\
\textbf{Speed} & ++& +& +++ & ++(PBR)/-(RTX) & + & + & +  \\
\textbf{Human Interface} 
&\begin{tabular}{@{}c@{}}Mouse \\ Keyboard  \end{tabular} 
& -
&\begin{tabular}{@{}c@{}}Mouse \\ Keyboard  \end{tabular} 
& - 
&\begin{tabular}{@{}c@{}}Mouse \\ Keyboard  \end{tabular} 
&\begin{tabular}{@{}c@{}}Natural \\ Language  \end{tabular} 
&\begin{tabular}{@{}c@{}}Virtual \\ Reality  \end{tabular}   \\
\begin{tabular}{@{}c@{}}\textbf{Integrated} \textbf{Motion Planner}\end{tabular}
 & Yes & No & No & No & No & No & No  \\
  \midrule
  \rowcolor{gray!0}
\textbf{Specialty} & 
\begin{tabular}{@{}c@{}}Phys. Int. in\\Large Scenes\end{tabular} &
\begin{tabular}{@{}c@{}}Nav.\end{tabular} & 
\begin{tabular}{@{}c@{}}Fast, \\ Nav. \end{tabular}& 
\begin{tabular}{@{}c@{}}Articulation,\\ Ray Tracing\end{tabular} & 
\begin{tabular}{@{}c@{}}Object States,\\ Task Planning \end{tabular} & 
\begin{tabular}{@{}c@{}} Object States,\\Task Planning \end{tabular} & 
\begin{tabular}{@{}c@{}}Audio,\\Fluids \end{tabular} \\

\bottomrule

\rowcolor{gray!0}
\multicolumn{8}{l}{\scriptsize \textbf{Type of rendering:} \textbf{PBR}:Physics-Based Rendering, \textbf{IBR}:Image-Based Rendering, \textbf{RTX}:Ray Tracing }\\
  \rowcolor{gray!0}
\multicolumn{8}{l}{\scriptsize \textbf{Virtual sensor signals:} \textbf{RGB}: Color Images, \textbf{D}:Depth, \textbf{N}:Normals, \textbf{SS}:Semantic Segmentation, \textbf{LiDAR}:Lidar, \textbf{FL}:Flow (optical and/or scene), \textbf{S}: Sounds}\\

\end{tabular}
\vspace{-3mm}
\end{table*}

The use of simulation in robotics has significantly increased in recent years and the research community has proposed a number of simulators for robotics and embodied AI.
Here, we use the terms physics simulator and simulation environment as follows.
A \textbf{physics simulator} is an engine capable of computing the physical effect of actions on an environment (e.g. motion of bodies when a force is applied, or flow of liquid particles when being poured)~\cite{coumans2013bullet,todorov2012mujoco,lee2018dart,unity,liang2018gpu,ode}. On the other hand, a \textbf{simulation environment} is a framework that includes a physics simulator, a renderer of virtual signals, and a set of assets (i.e. models of scenes, objects, robots) ready to be used to study and develop solutions for different tasks. Both components are crucial for advancing embodied AI and robotics. Here, we focus on the discussion of simulation environments.

Several simulation environments have been proposed recently to study manipulation with stationary arms~\cite{yu2020meta,james2020rlbench,zhu2020robosuite,erickson2020assistive,lee2019ikea}. 
Most of them are based on Bullet~\cite{coumans2013bullet} or MuJoCo~\cite{todorov2012mujoco} for physics simulation, and render images with the default renderer or a Unity~\cite{unity} plugin. Different from these simulation environments, iGibson focuses on large-scale (house-size) scenes and includes fifteen fully interactive scenes with 108 rooms -- such as kitchens and bedrooms -- where researchers can develop solutions for navigation, manipulation and mobile manipulation. 

Closer to iGibson are simulation environments that include large-scale realistic scenes (e.g. homes or offices), which we summarize and compare in Table~\ref{t:simenv}. Gibson~\cite{xia2018gibson} (now Gibson v1) was the precursor of iGibson. It includes over 1400 3D-reconstructed floors of homes and offices with real-world object distribution and layout. Although Gibson incorporates PyBullet as its physics engine for simulating robot navigation, each scene is one single fully rigid object (static mesh). Thus, it does not allow agents to interact with the scenes other than collisions with the mesh, restricting its use to only navigation. A similar environment is Habitat~\cite{habitat19arxiv}. Despite its high rendering speed, Habitat uses the non-interactive assets from Gibson v1~\cite{xia2018gibson} and Matterport~\cite{chang2017matterport3d} and therefore only supports navigation tasks and simplified rearrangement of added objects.
Recent work~\cite{xia2020interactive} introduced an extension to the Gibson v1 environment to support Interactive Navigation~\cite{xia2020interactive} where parts of the reconstructions corresponding to five object classes (chairs, tables, desks, sofas, and doors) in several Gibson static models were segmented and replaced with interactive versions. This enabled navigation agents to interact with the scene, and thus allowed for the first benchmark for Interactive Navigation. 
However, the simulators above are not interactive \cite{xia2018gibson,habitat19arxiv} or partially interactive 
\cite{xia2020interactive}.
iGibson encapsulates the functionalities in all previous versions of Gibson with backwards compatibility for the static environments.

A variety of simulation environments have been proposed recently for scene-level interactive tasks, such as Sapien~\cite{xiang2020sapien}, AI2Thor~\cite{kolve2017ai2}, VirtualHome~\cite{puig2018virtualhome}, and ThreeDWorld (TDW~\cite{gan2020threedworld}). 
These simulators adopt different ways of agent-world interactions. \textbf{Predefined  Actions (PA)} consist in the set of actions that can be performed for each object type. When the agent is close enough to an object and the object is in the right state (precondition), the agent can select a predefined action, and the object is ``transitioned'' to the next state (postcondition). In Table~\ref{t:simenv} we refer to this technique as Rigid Bodies with Predefined Actions (RBPA). 
It is possible to combine Rigid Bodies with Predefined Actions (RBPA) and Rigid Body Physics (RBP), such as first using RBPA to grasp an object and then using RBP after releasing it.
AI2Thor and VirtualHome use predefined actions (PA) as an abstraction of physical interactions and allow agents to modify the object states instantaneously (e.g. open a closed cabinet), suitable to study high-level task planning.
While the availability of PA helps focusing on symbolic reasoning, the full robotics problems require RBP to simulate the tasks in all its complexity. PA limits access to all granularities of the task, which impedes robot learning and sim2real transfer.
With the purpose of simulating full robotics tasks, iGibson uses RBP, simulating the physics behavior of all objects continuously, with embodiment of real robots. This is crucial if the learned policy is to be deployed in the real world.
Similar to iGibson, Sapiens also uses RBP without PA, but with smaller scenes, focusing on interaction with articulated objects. In iGibson, we enrich the articulated objects models from the PartNet-Mobility dataset introduced by Sapien with materials and dynamics properties.
TDW is also capable of continuous RBP, with simplifications that facilitate grasping using robot avatars and not real robot platforms.

\begin{figure}[t]
\centering
    \begin{subfigure}[b]{0.145\textwidth}
        \centering
    \includegraphics[width=\textwidth]{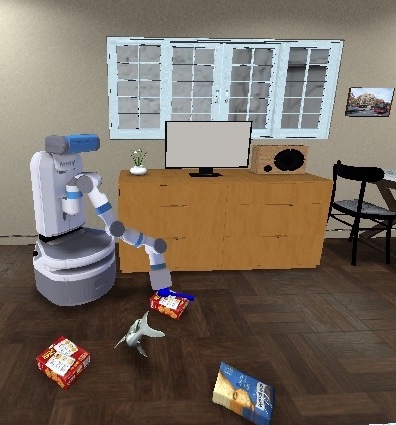}
    \caption{pick}
    \end{subfigure}
    \begin{subfigure}[b]{0.151\textwidth}
        \centering
    \includegraphics[width=\textwidth]{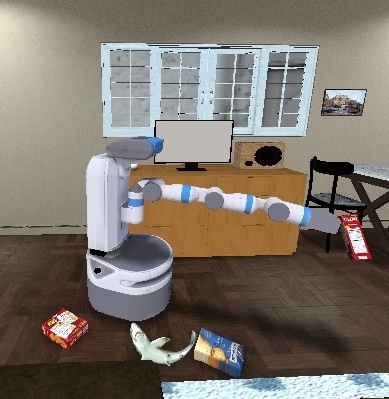}
    \caption{transport}
    \end{subfigure}
    \begin{subfigure}[b]{0.153\textwidth}
        \centering
    \includegraphics[width=\textwidth]{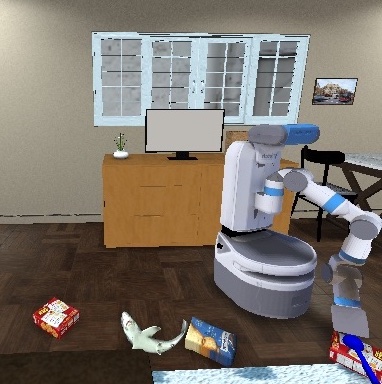}
    \caption{place}
    \end{subfigure}
    
    \caption{Simulated Fetch Robot performing a mobile manipulation task in a cluttered environment. 
    }
    \label{fig:mobile_manip}
\vspace{-5mm}
\end{figure}

While other simulation environments focus on particular aspects of embodied agent simulation (Table.~\ref{t:simenv}), iGibson \textbf{uniquely} unifies a set of important tools for robotics that together enable robot learning in large scenes: support of LiDAR and PBR rendering, speed that enable reinforcement learning, robot integration (URDF, controllers, motion planners), and continuous physics simulation of agents and objects.
This integration of features enables tasks such as mobile manipulation, illustrated in Fig. \ref{fig:mobile_manip}, including physics-based robot simulation across the entire task.

\section{iGibson Simulation Environment}
\label{s_feats}

In this section, we discuss the main structure, properties and features of iGibson that support training of robust sensor-guided policies for navigation and manipulation.

\subsection{\textbf{Simulation Characteristics and API}}

At the highest level, iGibson follows the OpenAI Gym~\cite{brockman2016openai} convention. The environment receives an action and returns a new observation, reward and additional meta-information (e.g. if the episode has ended). Environments are specified with config files that determine scenes, tasks, robot embodiments, sensors, etc. Given a config file, iGibson creates an \texttt{Environment} that contains a \texttt{Task} and a \texttt{Simulator}. The \texttt{Simulator} contains a \texttt{Scene}, with a list of interactive \texttt{Object}s and one or more \texttt{Robot} instances. It also contains a \texttt{Renderer} that generates virtual visual signals from any point of view, such as a camera mounted on a robot or an external third person view. The \texttt{Task} defines the reward, initial and final conditions for the scene and the agents. While modular and easy to extend, most users may only need to interface with \texttt{Environment} after instantiating it with the appropriate config files.

iGibson comes with multiple easy-to-use configs, demos and Docker~\cite{merkel2014docker} files. It has been extensively adopted 
to train visuo-motor policies that successfully transfer to the real world~\cite{meng2020scaling, meng2019neural, kang2019generalization, hirose2019deep}, and was the platform for iGibson Sim2Real Challenge at CVPR20~\cite{sim2real_challenge} and iGibson Challenge at CVPR21~\cite{igibson_challenge_2021}. The provided virtual LiDAR sensor has been used for robotics research in planning and reinforcement learning for social navigation~\cite{DArpino_ICRA2021} and mobile manipulation~\cite{xia2020relmogen}. iGibson is easily parallelizable and supports off-screen rendering on clusters.

\subsection{\textbf{Fully Interactive Assets}}
\begin{figure*}
    \centering
    \includegraphics[width=0.98\textwidth]{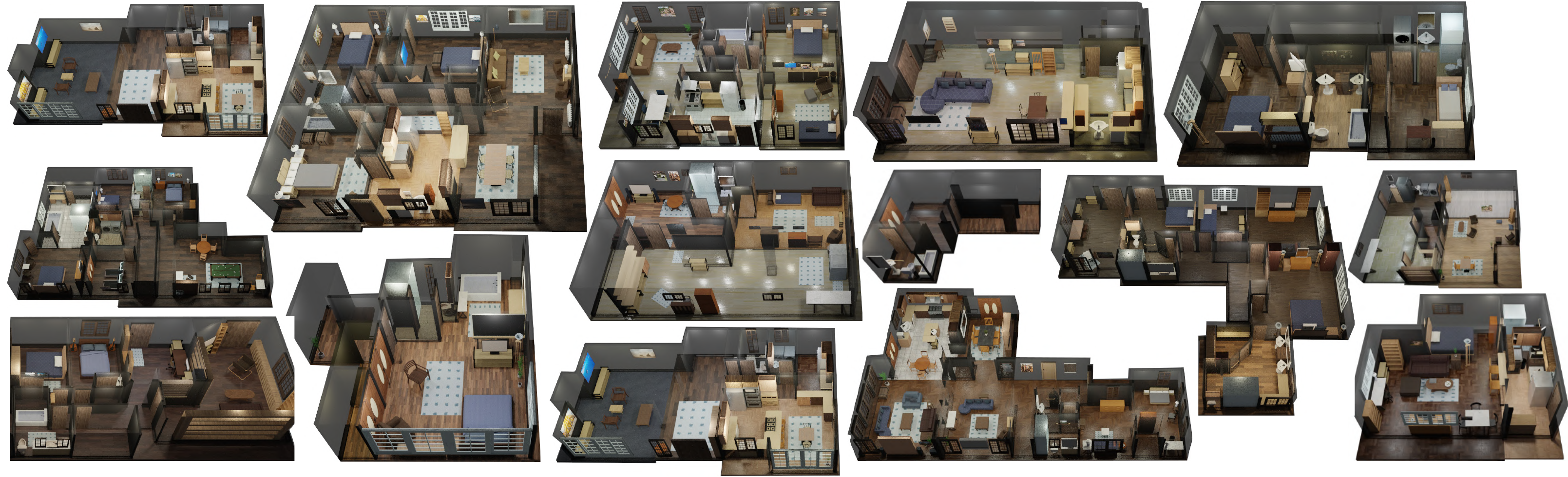}
    \caption{Fifteen interactive iGibson 1.0 scenes modelled after real-world reconstructions, preserving layout, distribution and size of objects.}
    \label{fig:scenes}
\vspace{-5mm}
\end{figure*}

iGibson provides fifteen high quality fully interactive scenes with 108 rooms (see Fig.~\ref{fig:scenes}), populated with interactable objects. The scenes are interactive versions of fifteen 3D reconstructed scenes included in the Gibson v1 dataset. To preserve the real-world layout and distribution of objects, we follow a semi-automatic annotation procedure. This process is radically different from the annotation we performed for the Interactive Gibson Benchmark~\cite{xia2020interactive}. Instead of segmenting the original scene and replacing part of the meshes with interactive object models, we create fully interactive counterparts of the 3D reconstruction from scratch. This eliminates the need to fix artifacts in the original mesh due to reconstruction noise or segmentation error, and allows us to improve the overall quality of the scenes.

The scene generation process is composed of two annotation phases. First, the layout of the scene is annotated with floors, walls, doors and window openings. Then, all objects are annotated with 3D bounding boxes and class labels. We annotate bounding boxes for 57 different object classes, including all furniture types (doors, chairs, tables, cabinets, TVs, shelves, stoves, sinks, etc) and some small objects (plants, laptops, speakers, etc); see project website for the complete list. Annotating class-labeled bounding boxes allows us to scale and use different models of the same object class, while maintaining the real-world distribution of objects in the scene. In this way, we are able generate realistic randomized versions of the scenes (see Sec.~\ref{ss_dr}). To achieve the highest quality, for each class-labeled bounding box, we select a best fitting object model. The scene is also annotated with lights, with which we generate light probes for physics-based rendering (see Sec.~\ref{ss_sensing}). We also bake in a realistic ray-traced ambient light and other light effects in the walls, floors and ceilings.

The object models are curated from open-source datasets: ShapeNet~\cite{chang2015shapenet}, PartNet Mobility~\cite{xiang2020sapien,mo2019partnet}, and SketchFab. To preserve visual realism of the original reconstruction, we improve the object visual quality by annotating different parts of the models with photorealistic materials, which are then used by iGibson's physics-based renderer. We utilize materials from CC0Texture, including wood, marble, metal, etc. To achieve a high degree of physics realism, we curate a mapping from visual materials to friction coefficients. We additionally compute the collision mesh, center of mass and inertia frame for each link of all objects. To assign realistic mass and density for different objects, we take the the median values of the top 20 search results from Amazon.

Additionally, we provide compatibility with CubiCasa5K~\cite{kalervo2019cubicasa5k} and 3D-Front~\cite{fu20203d} repositories of home scenes. We use their scene layouts and populate them with our annotated object models, leading to additional more than 12000 interactive home scenes. These scenes contain fewer objects than the fifteen iGibson scenes, but provide a very large number of additional models to train tasks. 

The fully interactive scenes we include in iGibson enable learning of interactive tasks in large realistic home scenes; in Sec.~\ref{ss_pre_exp} we show that the scenes can be used to learn a useful visual representation that accelerates the learning of downstream manipulation tasks.

\begin{figure}
    \centering
    \includegraphics[width=0.98\columnwidth]{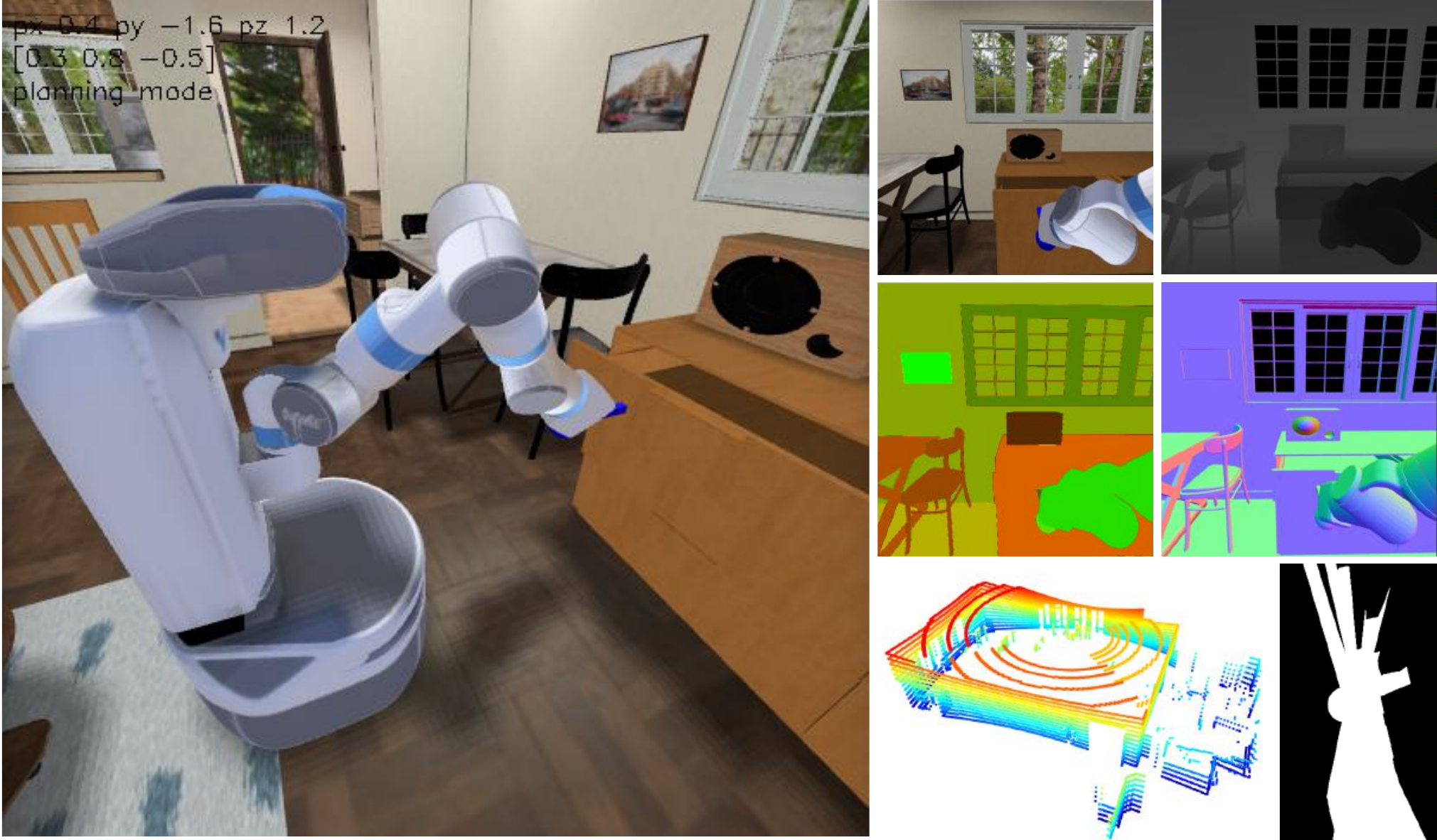}
        \vspace{-1mm}
    \caption{Robot interacting in iGibson 1.0 (large picture: 3rd person view) and \textbf{virtual sensor signals} generated. Policies and solutions can make use of the following channels: (from top to bottom, left to right) RGB, depth, semantic/instance segmentation, normals, 16D LiDAR (point cloud), 1D LiDAR (also as occupancy map). Not depicted: optical/scene flow, joint encoders for robot's and objects' joints, poses, wrenches, contact points, and map localization. \vspace{-5mm}}
    \label{fig:channels}
\end{figure}

\subsection{\textbf{Virtual Sensors}}
\label{ss_sensing}

A crucial component of iGibson is the generation of high quality virtual sensor signals, i.e. images and point clouds, for the simulated robots. In the following, we summarize the most relevant of these signal generators (Fig.~\ref{fig:channels}).

\noindent\subsubsection*{Physics Based Rendering}
In iGibson, we include an open-source physics-based renderer, which implements an approximation of BRDF models~\cite{schlick1994inexpensive} with spatially varying material maps including roughness, metallic and tangent-space surface normals, extending ~\cite{pbrgithub}.

\noindent\subsubsection*{LiDAR Sensing}
Many real-world robots are equipped with LiDAR sensors for obstacle detection. In iGibson, we support virtual LiDAR signals, with both 1 beam (e.g. Hokuyo) and 16 beams (e.g. Velodyne VLP-16). We include a simple drop-out sensor noise model to emulate the common failure case in real sensors in which some of the laser pulses do not return. Additionally, we provide the functionality to turn the 1D LiDAR scans into local occupancy maps, which are bird's-eye view images with three types of pixels indicating free, occupied, or unknown space.

\noindent\subsubsection*{Additional Visual Channels}
In addition to RGB and LiDAR, we support a wide range of visual modalities, such as depth maps, optical/scene flow and normals, segmentation of semantic class, instance, material and movable parts. These modalities can support research topics such as: depth/segmentation/normal/affordance prediction~\cite{porzi2016learning, casser2019depth, xie2020best}, action-conditioned flow prediction~\cite{nematollahi2020hindsight}, multi-modal pose estimation~\cite{choi2016rgb, wang20206, hu2019segmentation}, and visuomotor policy training assuming perfect vision systems~\cite{xia2020interactive, yan2020close}.

\subsection{\textbf{Domain Randomization}}
\label{ss_dr}

It is standard practice for robot learning to partially randomize the environment's parameters in order to make the policy more robust~\cite{sadeghi2016cad2rl,tobin2017domain,james2019sim,andrychowicz2020learning}. With the model being trained in a wide distribution of environments, it will be more likely to generalize to unknown evaluation environments. The evaluation environment may be the real world if we aim to train in simulation and transfer the policy to a real robot. In iGibson, we include domain randomization that leads to an endless variation of visual appearance, dynamics properties and object instances with the same scene layout.

First, we provide \textbf{object randomization}. 
The original 3D reconstructions are annotated with class-labeled object bounding boxes. These labels can be used to instantiate any object model of the corresponding class into the given bounding box (e.g. a bounding box labeled as ``table" can be filled with any table model). 
This randomization maintains the semantic layout of the scenes (i.e. the object categories remain at the same 3D locations) while enabling near-infinite combinations of object instances. It provides strong variation in depth maps and LiDAR signals that helps robustify policies based on these observations (see Sec.~\ref{ss_dr_lidar_exp}).

Second, we provide \textbf{material randomization}.
In addition to high-quality material annotation for object and scene models, we provide 
a mechanism to randomize the specific material model associated with each object part (e.g. associating a different type of wood or metal). 
The effect is a stark color randomization that still represents plausible material combinations. This randomization generates strong variations in the RGB images and helps robustify policies based on this observation (see Sec.~\ref{ss_dr_lidar_exp}). Moreover, the dynamics properties of all object links can be randomized based on a curated mapping from visual materials to dynamics properties.

\subsection{\textbf{Motion Planning}}
\label{ss_mp}

Motion planners provide collision-free trajectories to move a robot from an initial to a final configuration~\cite{lavalle2006planning}. 
They can be used to generate collision-free navigation paths for robot bases and collision-free motion paths for robot arms.
In iGibson, we include implementations of the most popular sampling-based motion planners: rapidly growing random trees (RRT~\cite{lavalle1998rapidly}) and its bidirectional variant (BiRRT~\cite{qureshi2015intelligent}), and lazy probabilistic road-maps (lazyPRM~\cite{bohlin2000path}), adapted from~\cite{sspybullet}.
Sampling-based motion planners can have rather suboptimal and intricate paths. To alleviate this, we include acceleration-bounded shortcuts~\cite{hauser2010fast} for smoother paths.

\subsection{\textbf{Human-iGibson Interface}}
\label{ss_hii}
We provide a human-iGibson interface that enables 
users to navigate and interact in iGibson scenes using 
mouse and key commands on a viewer window. 
The user can navigate and interact with (pull, push, pick and place) objects. 
While a virtual reality or a 3D mouse interface may provide a more intuitive experience, most users do not have the necessary hardware.
This interface offers a natural and simple way to demonstrations for imitation learning, evaluate the difficulty or feasibility of a task, or change the scene into a better initial state, for example. 
This interface is also integrated with the motion planner to command the robot to desired base and/or arm configurations.
We verify this interface facilitates efficient development of interactive robotic solutions in Sec.~\ref{ss_il} \vspace{-3mm}.

\section{Experiments}
\label{s_exp}

The goal of our experiments is to study how iGibson's features help to develop AI agents. Specifically, we examine:
\begin{itemize}[wide, labelwidth=!, labelindent=0pt]
\item (Sec.\ref{ss_dr_lidar_exp}) does iGibson's \textbf{domain randomization} and realistic virtual sensor signals (including \textbf{LiDAR}) allow navigation agents to generalize to unseen scenes (including the real world)?
\item (Sec.\ref{ss_il}) can the \textbf{human-iGibson interface} be used to efficiently train imitation learning agents for manipulation and \textbf{mobile manipulation tasks}?
\item (Sec.\ref{ss_pre_exp}) does the \textbf{full interactivity} in the scenes allow agents to learn visual representations that accelerate learning of downstream manipulation tasks? 
\end{itemize}

\subsection{\textbf{Domain Randomization and Realistic Virtual Sensor Signals for Robot Navigation}}
\label{ss_dr_lidar_exp}

In the first three experiments, we evaluate the generalization benefits brought by our realistic virtual sensor signals (including LiDAR) and domain randomization. First, we compare the generalization capabilities of vision-based reinforcement learning policies trained with and without domain randomization. Concretely, we evaluate the performance of policies trained in iGibson for PointGoal tasks~\cite{anderson2018evaluation} in held-out scenes with held-out visual textures. The observations for the policy include the depth maps, the robot's linear and angular velocities, the goal location in the robot's reference frame, and the next 10 waypoints in the shortest path between the robot's current location and the goal location, separated by \SI{0.2}{\meter}. The waypoints are computed only based on the room layout, not the objects, so the robot mainly relies on depth maps for obstacle avoidance.

Second, we evaluate the performance of robot policies trained in iGibson to navigate to a target object (a lamp) using virtual RGB images also in held-out scenes with held-out visual textures. The task goal is to get at least 5\% of the image occupied by the pixels of the target object. The observation for the policy only includes RGB images. In the above two experiments, we train in eleven scenes and evaluate in four held-out scenes with held-out visual textures.  

\begin{figure}
    \centering
    \includegraphics[width=0.85\columnwidth]{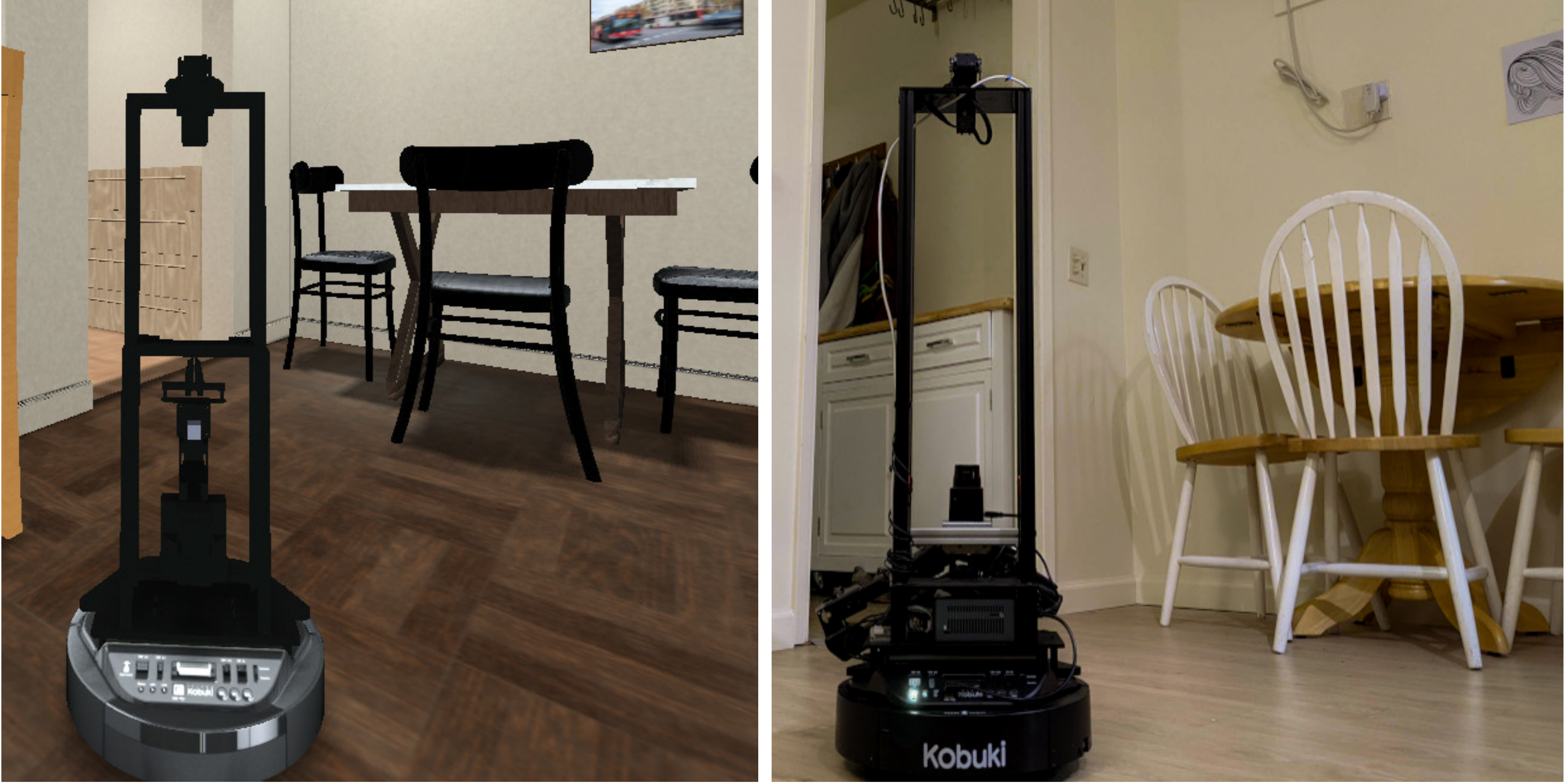}
    \caption{Robot navigating the real-world counterpart of the iGibson 1.0 scene \texttt{Rs\_int}. The robot executes a policy trained in simulation with virtual LiDAR signals, without domain adaptation. The quality and realism of iGibson facilitates zero-shot policy transfer. \vspace{-5mm}}
    \label{fig:exp3}
\end{figure}

Third, we evaluate the performance of the policies trained in iGibson for a PointGoal tasks~\cite{anderson2018evaluation} using \textbf{virtual LiDAR signals} with no vision inputs, and examine how well those policies transfer to the real world without adaptation, a hard test for generalization. The observations for the policy include a 1D LiDAR scan with 512 laser rays, the robot's linear and angular velocities, and the goal location in the robot's reference frame. To focus on sim2real transferability, we train in our scene \texttt{Rs\_int}, for which we have access to the real-world counterpart (see Fig.~\ref{fig:exp3}).

\paragraph*{Results} In the first two experiments, we observe better generalization capabilities in policies using iGibson's domain randomization. For PointGoal navigation based on depth images, the performance goes from 0.27 to 0.40 SPL~\cite{anderson2018evaluation} and from 31.25\% to 44.75\% success rate when using randomization, indicating that the larger variety of shapes observed in the training process generates more robust depth-based policies. For object navigation based on RGB images, the performance goes from 49.75\% to 57.5\% success rate, indicating that material randomization helps in obtaining RGB-based policies that are more generalizable to unseen scenes and textures. Finally, for PointGoal navigation based on LiDAR signals, the policy achieves 33\% success rate in \texttt{Rs\_int} in iGibson, and 24\% success rate in the real-world apartment. With only a 9\% drop in performance and the failures mostly occurring in the same episodes (same pairs of the initial and goal locations in iGibson and real world), this experiment indicates that the LiDAR signals generated in iGibson are realistic enough to facilitate zero-shot policy transfer. In summary, as shown in Table.~\ref{t:simenv}, iGibson provides unique support to train with realistic virtual sensor signals (e.g. LiDAR) and domain randomization, which leads to more robust robot navigation policies that successfully transfer to novel scenes.

\begin{figure}
\centering
\begin{subfigure}[b]{0.48\textwidth}
 \centering
 \includegraphics[width=0.32\textwidth]{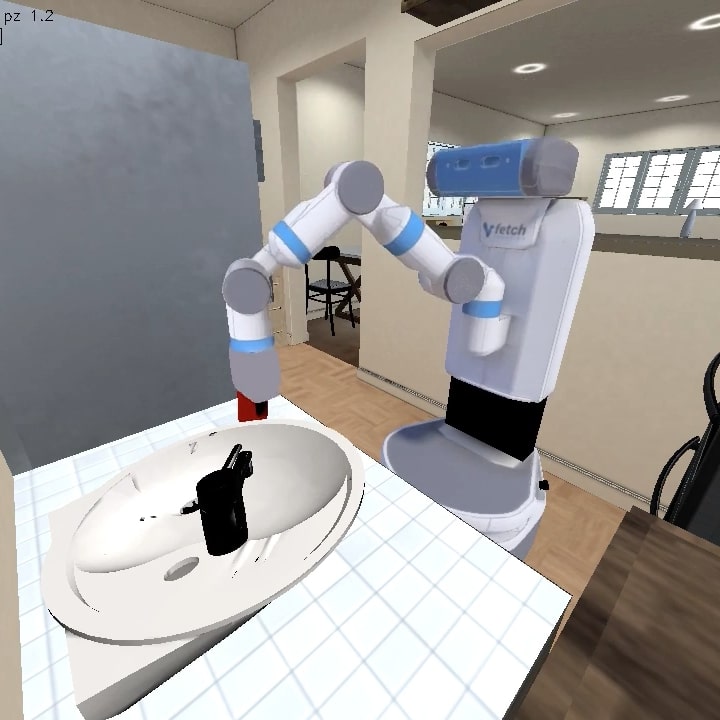}
 \includegraphics[width=0.32\textwidth]{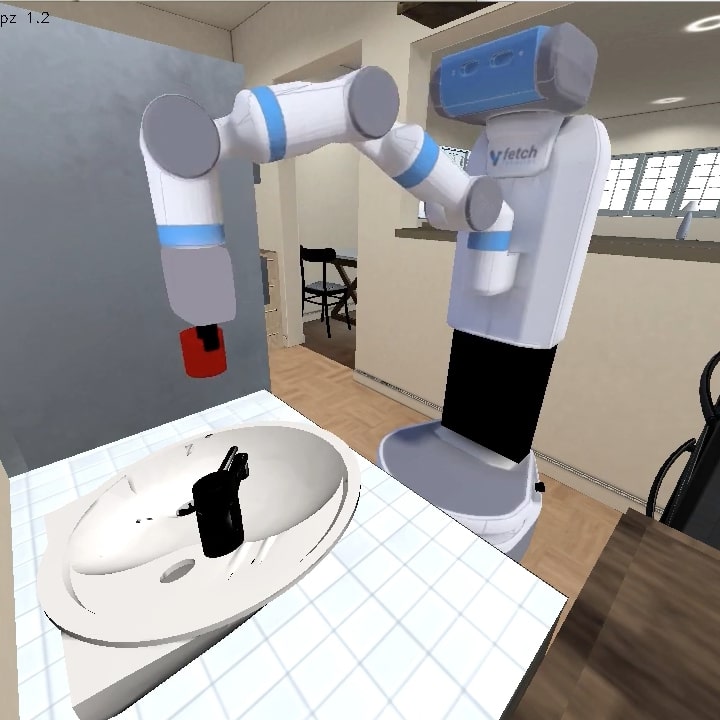}
 \includegraphics[width=0.32\textwidth]{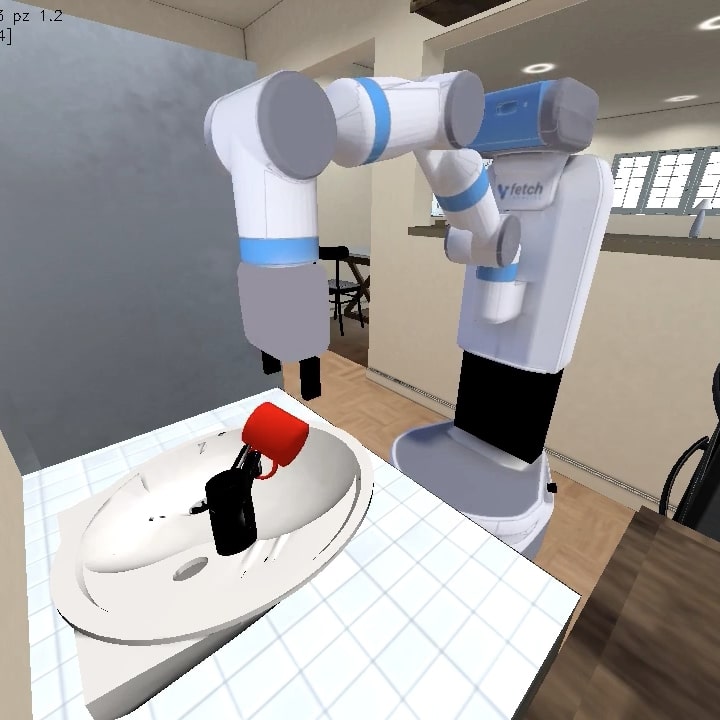}
 \\
 \vspace{4pt}
 \includegraphics[width=0.32\textwidth]{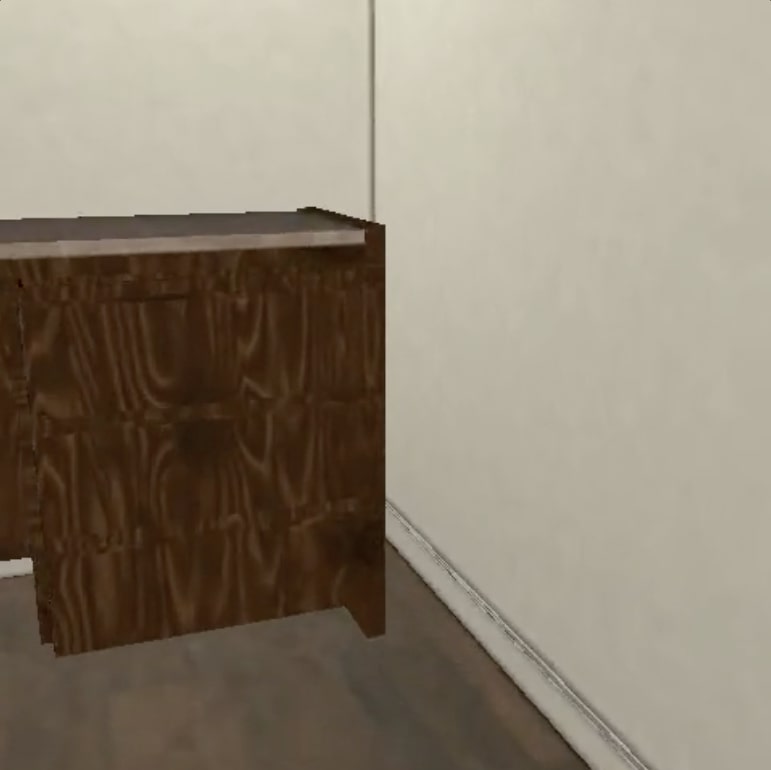}
 \includegraphics[width=0.32\textwidth]{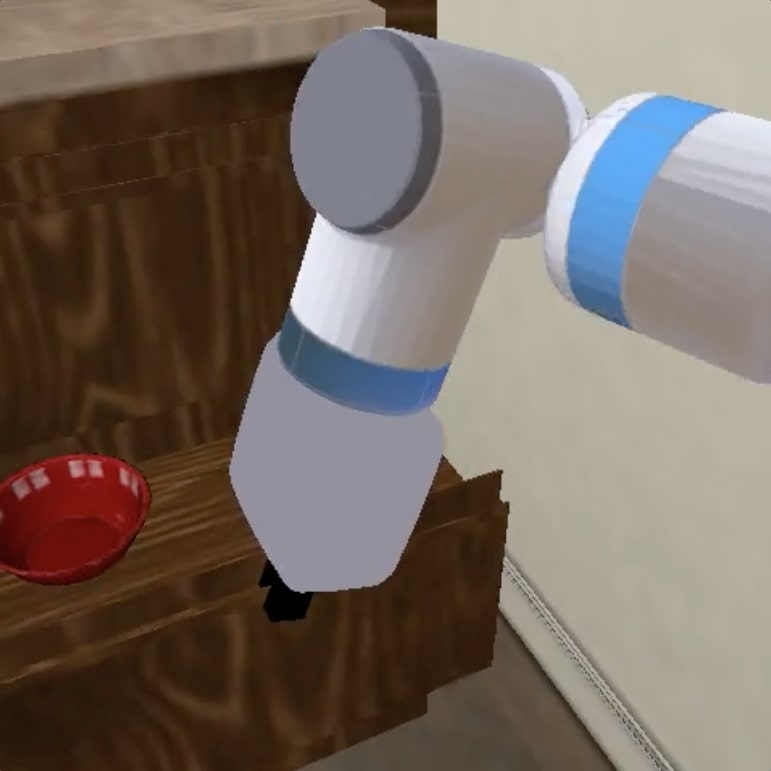}
 \includegraphics[width=0.32\textwidth]{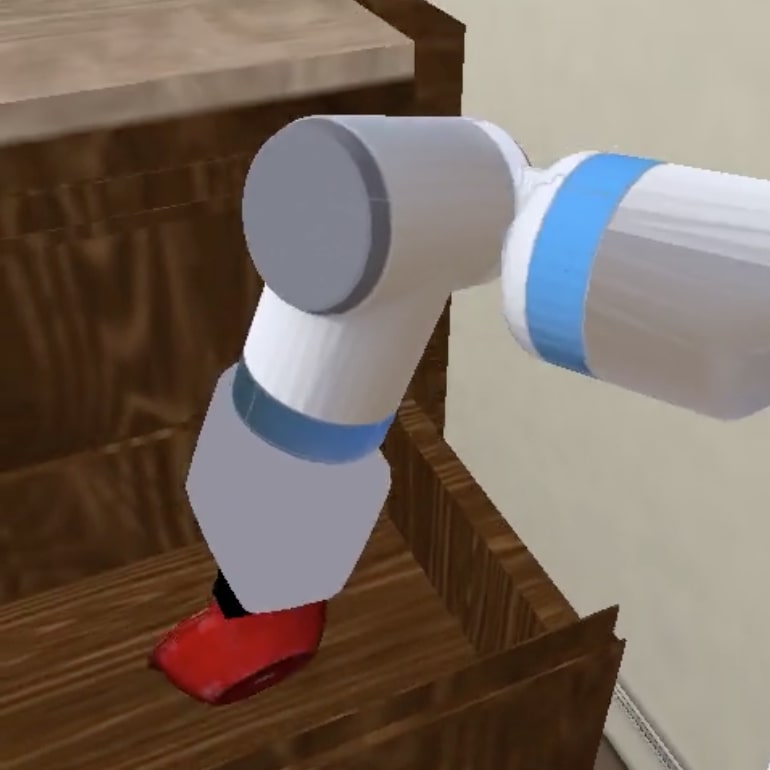}
\end{subfigure}
\caption{{Imitation learning from human demonstration. \textit{Top:} third-person view of a robot performing a pick-and-place task. The policy trained using demonstrations collected with our human-iGibson interface achieves 98\% success rate. \textit{Bottom:} first-person view of a robot performing a mobile manipulation task. The policy trained using teleoperated demonstrations achieved 70\% success rate.} \vspace{-5mm}}
\label{fig:exp4}
\end{figure}

\subsection{\textbf{Imitation Learning of Human Demonstrated (Mobile) Manipulation}}
\label{ss_il}

In the second set of experiments, we evaluate iGibson as platform to train robots to perform manipulation and mobile manipulation tasks with Imitation Learning. First, we test the usability of the human-iGibson interface to efficiently collect demonstrations of \textbf{manipulation-only} tasks. We collect 50 demonstrations of pick-and-place operations with 20 mug models: pick a mug and place in the sink (Fig.~\ref{fig:exp4}), and store pairs of state (object position) and action (desired end-effector translation).
We use the demonstrations to train a behavioral cloning policy that maps states to actions at \SI{20}{\hertz}. The action space consists of two parts: a 3-dimensional continuous space for desired end-effector translation, and a 1-dimensional discrete space for gripper opening. The desired end-effector translation is computed using inverse kinematics (IK) and executed with a joint position controller. The evaluation is conducted using a simulated mobile manipulator (Fetch robot), and generalization is tested with 5 unseen mugs.

Second, we collect demonstrations for imitation learning through continuous teleoperation using a system similar to Roboturk~\cite{mandlekar2018roboturk} for \textbf{mobile-manipulation} tasks. We collect 200 demonstrations with a simulated Fetch robot for search-and-pick operations: the robot must navigate and interact with a cabinet to open drawers, find a bowl and pick it (Fig.~\ref{fig:exp4} (bottom)). The bowl and robot initial poses are randomized between episodes. Using these demonstrations, we train a behavioral cloning policy mapping observations to actions at \SI{20}{\hertz}. The observation space includes the robot proprioceptive information (joint positions and velocities) and RGB-D images from virtual cameras on robot's head and wrist. The action space consists of four parts: a 1-dimensional discrete value indicating whether to extend the arm, 2-dimensional continuous values representing the robot base's linear and angular velocities, 6-dimensional continuous values representing the desired pose change of the end-effector, and 1-dimensional discrete value indicating whether to open the gripper.

\paragraph*{Results}
In our first experiment, training manipulation-only policies with imitation learning, we observe 98$\%$ success rate over 100 evaluation episodes. This experiment showcases that the human-iGibson interface enables easy collection of effective demonstrations for imitation learning, and the integrated motion planner is helpful for policy training. These two integrated features, as shown in Table.~\ref{t:simenv}, are novel combinations offered by iGibson.
In the second experiment, training mobile-manipulation policies with imitation learning, we observe 70$\%$ success rate over 20 evaluation episodes. This experiment indicates that the we can leverage iGibson to train imitation learning algorithms for mobile manipulation tasks.

\subsection{\textbf{Pretraining in Fully Interactive Scenes}}
\label{ss_pre_exp}

In the third and final set of experiments, we evaluate the potential of using iGibson's fully interactive scenes to learn an intermediate visual representation that encodes the expected outcome of interactions with different objects. Such an intermediate visual representation may be used to accelerate robot learning of manipulation tasks, since they typically require the agent to associate visual observations with promising areas of interaction to change the state of the scene towards a manipulation goal.

\begin{figure}
    \vspace{2mm}
    \centering
    \begin{subfigure}{0.23\textwidth}
    \includegraphics[width=\columnwidth]{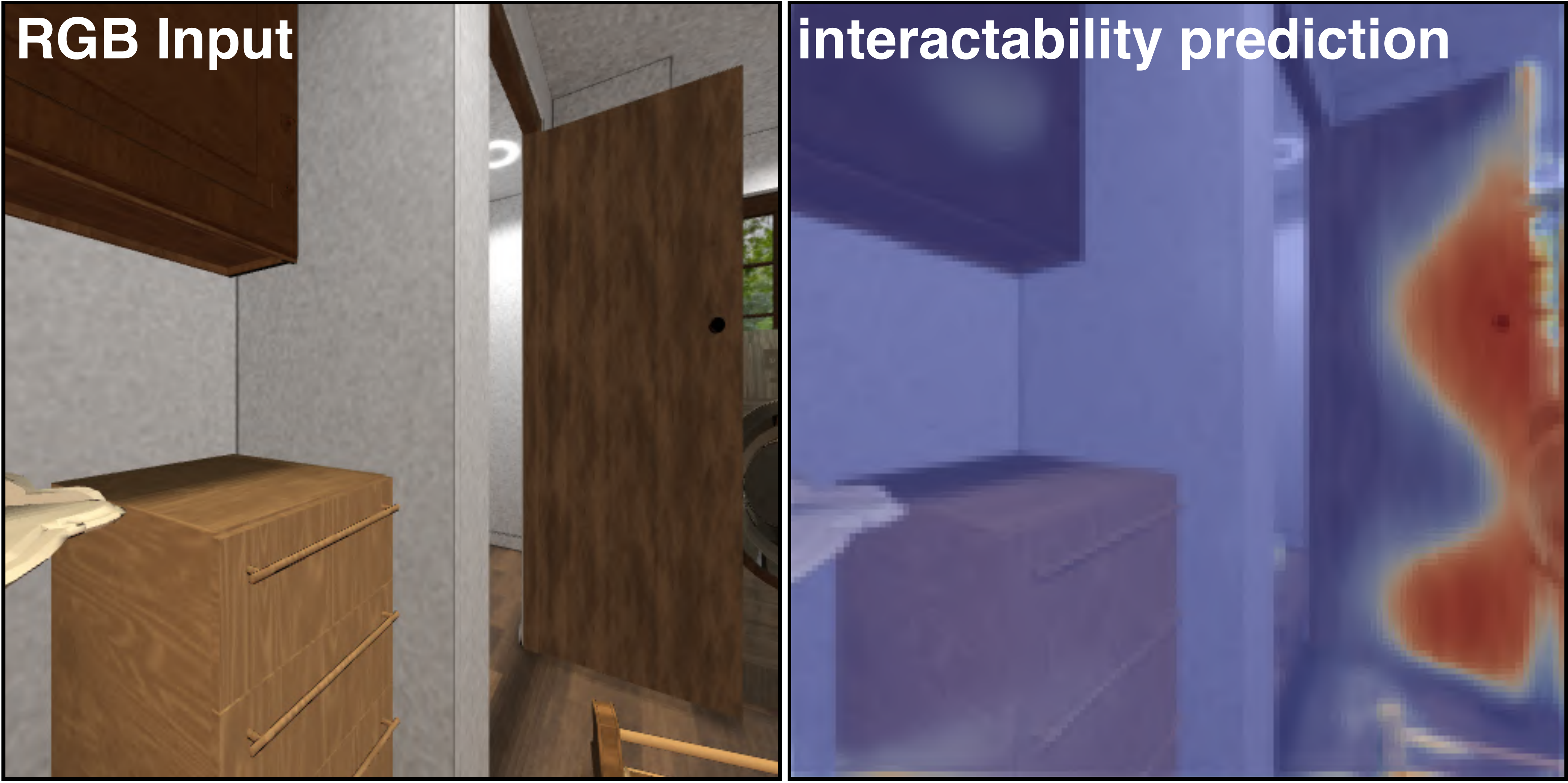}
    \end{subfigure}
    \begin{subfigure}{0.24\textwidth}
    \includegraphics[width=0.48\columnwidth]{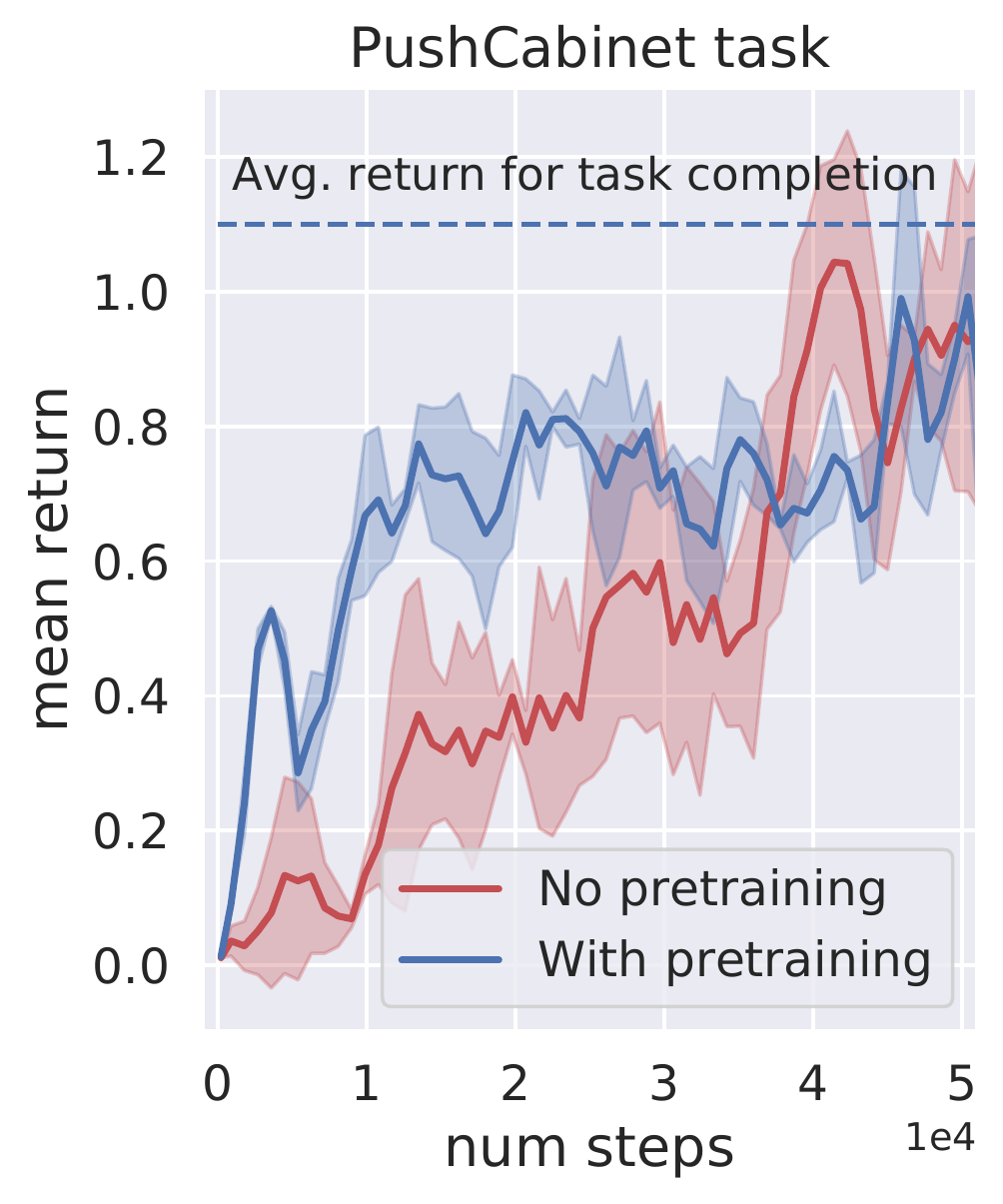}
    \includegraphics[width=0.48\columnwidth]{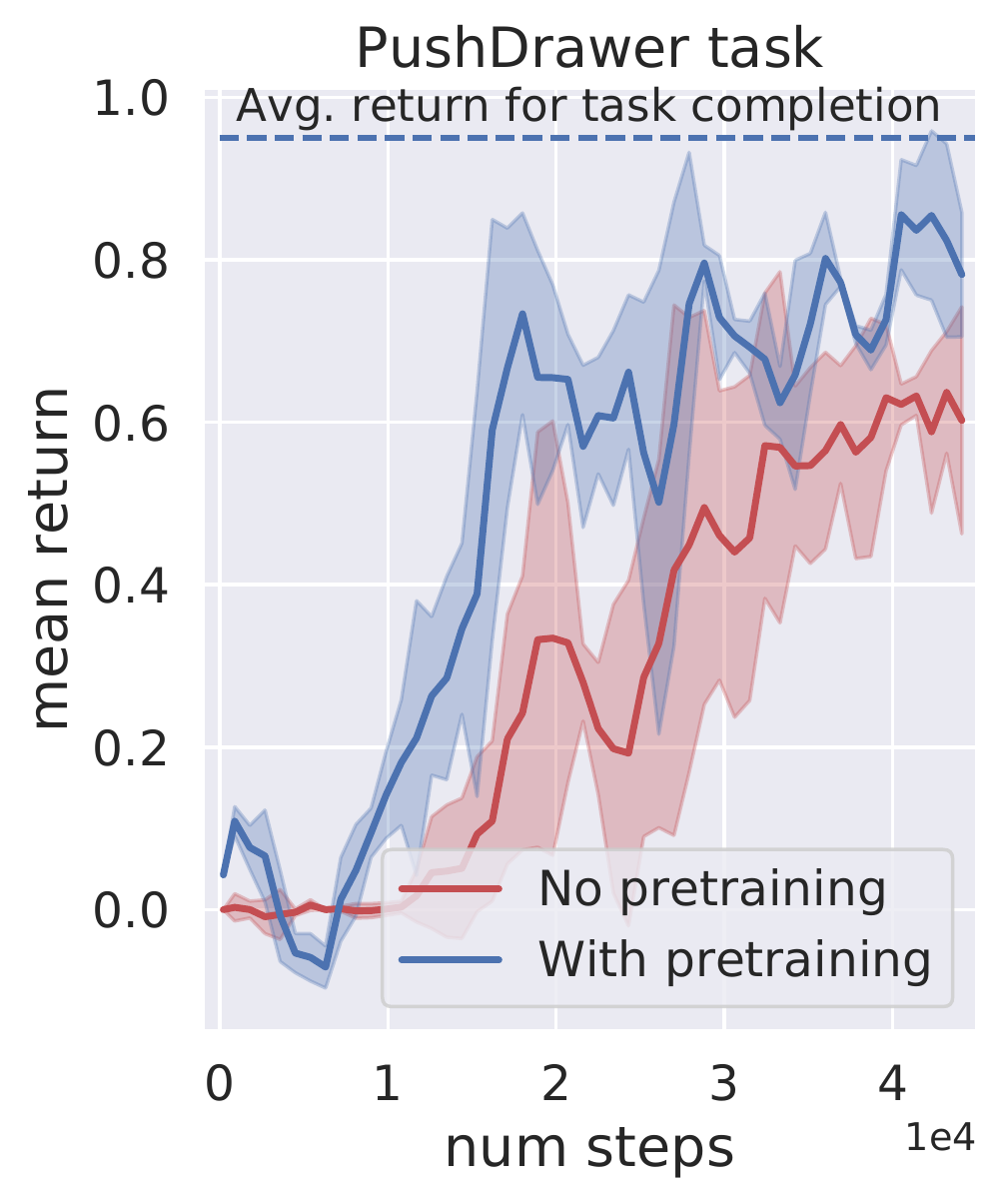}
    \end{subfigure}
        \vspace{-1mm}
    \caption{\textit{Left:} Example result of interaction pretraining. The agent receives RGB input and predicts if the pixels are pushable (red: higher probability; blue: lower) (Sec.~\ref{ss_pre_exp}). The model learns to associate edge of doors as the most pushable points. \textit{Right:} Training curves for two interactive tasks (\texttt{PushDrawer,PushCabinet}) with and without interaction pretraining. \vspace{-5mm}}
    \label{fig:interaction_pretrain}
\end{figure}

To learn such representations, we set up a virtual agent that interacts with random points in the scenes and learns to predict the outcome of these interactions. The interaction is parameterized as a coordinate in the virtual agent's image observation space. We emulate a pushing interaction by displacing the corresponding 3D location of the selected pixel by \SI{30}{\centi\meter} in the opposite direction of the surface normal, applying a maximum force of \SI{60}{\newton} (a common payload of commercial robots). A motion of the point for more than \SI{10}{\centi\meter} is considered a success. We sample 10 random pushes at each location, 4,000 locations in each scene. We use the images annotated with interaction successes/failures to train a U-Net~\cite{ronneberger2015u}-based visual encoder that predicts heatmaps of expected interaction success from RGB input.

For the second phase, we train two policy networks for two manipulation task respectively (\texttt{PushDrawer, PushCabinet}). The goal is to close the drawers or the cabinets. The policy outputs points to interact (push) that are given to one of our integrated motion planners to generate an arm motion~\cite{xia2020relmogen}. We use DQN~\cite{van2016deep} as policy learning algorithm. The predicted interaction heatmaps are used to gate the Q-value maps predicted by the network.

\paragraph*{Results}

Fig.~\ref{fig:interaction_pretrain} (left) depicts the result of the pretrained visual model. We observe that the heatmap has stronger activation at the edge of the door than in the area closer to the hinge, and weak activation on closed cabinets. The model learns to identify the best areas to push to cause motion (further visualizations on project website). For both downstream tasks, we observe that using the pre-trained representation significantly accelerates training (Fig.~\ref{fig:interaction_pretrain} (right)). This suggests that the full interactability of iGibson can help agents learn useful visual representation for downstream mobile manipulation tasks. Having a subset of objects not physically interactable will lead to false negatives during pretraining, and prevents successful representation learning. 
As shown in Table.~\ref{t:simenv} and discussed in Sec.~\ref{s_rw}, fully interactive scenes with continuous robot actions is a specialty of iGibson.

\vspace{-2mm}
\section{Conclusion}
\label{s_con}
We presented iGibson, a novel simulation environment for developing interactive robotic agents in large-scale realistic scenes. iGibson includes 15 fully interactive scenes with 108 rooms, and novel capabilities to generate high-quality virtual sensor signals, domain randomization, integration with motion planners, and human-iGibson interface. 
Through experiments, we showcased that iGibson helps to develop robust policies for navigation and manipulation. We hope that iGibson can aid researchers in solving complex robotics problems in large-scale realistic scenes.

\section{Acknowledgement}
\small{
We thank NVIDIA, Google, ONR MURI (N00014-14-1-0671), ONR (1165419-10-TDAUZ), Panasonic (1192707-1-GWMSX), Qualcomm and Samsung for their support.}

\renewcommand*{\bibfont}{\footnotesize}
 
\bibliographystyle{IEEEtranN}
\bibliography{bibliography}

\section*{Appendix}
\label{a_appendix}

\subsection{Details on the Experimental Evaluation}

In this section, we provide additional information on the experiments presented in Sec.~\ref{s_exp} such as training processes and architectures, or system characteristics.

\subsubsection{Domain Randomization for Visual Navigation}
\label{a_ss_drpbr_exp}
For PointGoal navigation, as mentioned in the main paper, the observations for the policy include the depth maps, the robot’s linear and angular velocities, the goal location in the robot’s reference frame, and the location of the next 10 waypoints in the shortest path between the robot’s  current location and the goal location. We concatenate the robot’s linear and angular velocities, the goal location, and the location of the next 10 waypoints together and denote them by sensor observations. For both depth maps and sensor observations, we adopt a frame stack of 4. The encoder of the policy network consists of 3 parts: (a) a 9-layer ResNet with Weight Standardization~\cite{qiao2019weight} and GroupNorm, followed by a 3-layer Conv1D block to encode depth maps; (b) a 2-layer MLP to encode sensor observations; (c) a 2-layer fusion MLP to encode the concatenation of depth maps embedding and sensor observations embedding. The learning rate for SAC is 1e-4. For object navigation, the observation only includes a frame stack of 4 RGB images. The encoder of the policy network is a 9-layer ResNet with Weight Standardization and GroupNorm, followed by a 3-layer Conv1D block to encode RGB images. The learning rate for SAC is 5e-4.
To accelerate the training of our visual policies, we implement a multi-GPU distributed reinforcement learning pipeline based on SAC~\cite{haarnoja2018soft}. iGibson can be easily parallelized and deployed on large computing clusters, which makes the training highly efficient. 

Tables~\ref{t:pointgoal_nav} and \ref{t:object_nav} include a breakdown analysis of the results of the experiment for different scenes and phases of the training process.

\begin{table*}[h!]
  \caption{Quantitative results on PointGoal navigation.}
  \label{t:pointgoal_nav}
  \centering
  \begin{tabular}{l|l|cc|cc|cc|cc|cc}
    \toprule
    \multicolumn{1}{l}{\multirow{2}{*}{Domain}} & \multicolumn{1}{l}{\multirow{2}{*}{Training}} & \multicolumn{2}{c}{\texttt{Rs\_int}} & \multicolumn{2}{c}{\texttt{Beechwood\_0\_int}} & \multicolumn{2}{c}{\texttt{Merom\_0\_int}} & \multicolumn{2}{c}{\texttt{Pomaria\_0\_int}} & \multicolumn{2}{c}{Overall} \\
    \cmidrule(lr{0.5em}){3-4}\cmidrule(lr{0.5em}){5-6}\cmidrule(lr{0.5em}){7-8}\cmidrule(lr{0.5em}){9-10}\cmidrule(lr{0.5em}){11-12}
    \multicolumn{1}{l}{Randomization} & \multicolumn{1}{l}{Steps} & \multicolumn{1}{c}{SPL} & \multicolumn{1}{c}{SR} & \multicolumn{1}{c}{SPL} & \multicolumn{1}{c}{SR} & \multicolumn{1}{c}{SPL} & \multicolumn{1}{c}{SR} & \multicolumn{1}{c}{SPL} & \multicolumn{1}{c}{SR} & \multicolumn{1}{c}{SPL} & \multicolumn{1}{c}{SR} \\
    \midrule
    w/o & 400k & 0.41 & 42\% & 0.21 & 21\% & 0.22 & 22\%  & 0.27 & 27\%  & 0.28 & 28\% \\ 
    w/ & 400k & 0.35 & 36\% & 0.25 & 26\% & 0.32 & 33\%  & 0.25 & 25\%  & 0.29 & 30\% \\ 
    w/o & 800k & 0.50 & 56\% & 0.14 & 17\% & 0.22 & 25\%  & 0.23 & 27\%  & 0.27 & 31.15\% \\ 
    w/ & 800k & 0.60 & 67\% & 0.37 & 40\% & 0.29 & 33\%  & 0.36 & 39\%  & 0.40 & 44.75\% \\ 
    \bottomrule
  \end{tabular}
\end{table*}

\begin{table*}[h!]
  \caption{Quantitative results on object navigation.}
  \label{t:object_nav}
  \centering
  \begin{tabular}{l|l|c|c|c|c|c}
    \toprule
    \multicolumn{1}{l}{\multirow{2}{*}{Domain}} & \multicolumn{1}{l}{\multirow{2}{*}{Training}} & \multicolumn{1}{c}{\texttt{Rs\_int}} & \multicolumn{1}{c}{\texttt{Beechwood\_0\_int}} & \multicolumn{1}{c}{\texttt{Merom\_0\_int}} & \multicolumn{1}{c}{\texttt{Pomaria\_0\_int}} & \multicolumn{1}{c}{Overall} \\
    \cmidrule(lr{0.25em}){3-3}\cmidrule(lr{0.25em}){4-4}\cmidrule(lr{0.25em}){5-5}\cmidrule(lr{0.25em}){6-6}\cmidrule(lr{0.25em}){7-7}
    \multicolumn{1}{c}{Randomization} & \multicolumn{1}{l}{Steps} & \multicolumn{1}{c}{SR} & \multicolumn{1}{c}{SR} & \multicolumn{1}{c}{SR} & \multicolumn{1}{c}{SR} & \multicolumn{1}{c}{SR}\\
    \midrule
    w/o & 100k & 43\% & 34\% & 47\% & 45\% & 42.25\% \\
    w/ & 100k & 40\% & 45\% & 52\% & 51\% & 47\% \\
    w/o & 200k & 50\% & 46\% & 48\% & 55\% & 49.75\% \\
    w/ & 200k & 55\% & 54\% & 64\% & 57\% & 57.5\% \\
    \bottomrule
  \end{tabular}
\end{table*}

\subsubsection{LiDAR-Based Point-to-Point Navigation}
\label{a_ss_p2p}
As mentioned in the main paper, the observations for the policy include a 1D LiDAR scan with 512 laser rays, the robot’s linear and angular velocities, and the goal location in the robot’s reference frame. Similar to PointGoal navigation in Sec.~\ref{a_ss_drpbr_exp}, we concatenate the robot’s linear velocity, angular velocity and the goal location and denote them by sensor observations. The controller runs at \SI{10}{\hertz}. For both LiDAR scans and sensor observations, we adopt a frame stack of 8. The encoder of the policy network consists of 3 parts: (a) a 3-layer MLP to encode LiDAR scans; (b) a 3-layer MLP to encode sensor observations; (c) a 3-layer fusion MLP to encode the concatenation of the flattened (along temporal dimension) LiDAR scans embedding and sensor observations embedding. The learning rate for SAC is 1e-4. 

As real robot platform we use a \href{http://www.locobot.org/}{Locobot}, a non-holonomic mobile base, with an additionally mounted Hokuyo 1 beam LiDAR sensor (see Fig.~\ref{fig:exp3}, left). The simulated agents matches the characteristics of the real robot and sensor. The robot is controlled via ROS~\cite{quigley2009ros}. Processing the images requires more computation than the one provided by the onboard computer of the Locobot; thus, we send the images to a desktop computer that hosts the policy and generate commands that are sent back to the robot. We use the system we developed for the CVPR Challenge ``Sim2Real with iGibson''~\cite{sim2real_challenge}.

\subsubsection{Imitation Learning: Human Demonstrated Manipulation}
\label{a_ss_il}
We collected 50 demonstrations of pick-and-place operations demonstrated with our Human-iGibson interface to pick a mug and place in the sink (Fig.~\ref{fig:exp4}, right) with 20 different mug models. We split the demos into train/val/test set with 35, 10, 5 demos and 1095, 304, 157 pairs of state and action, respectively. The state includes the object position and the action includes the desired delta translation to add to the object position in the next step and whether to open gripper. The starting position of the mug is randomized within a \SI{50}{\centi\meter} by \SI{10}{\centi\meter} area.

We trained a policy using behavioral cloning to map states to actions. The policy network is composed by three MLP layers and ReLU activation. We used MSE loss for the delta translation action and cross entropy loss for the binary gripper action. We used ADAM optimizer~\cite{kingma2014adam} with learning rate 0.1 and trained the policy for 1000 epochs after validation loss plateaus.

For evaluation, we deployed the policy on the simulated Fetch robot and achieved 98\% success rate for 100 evaluation episodes. The two failed episodes are caused by the robot prematurely opening its gripper and the mug being dropped outside the sink. The robot has a time budget of 500 action steps (25 seconds) to accomplish the task.

\subsubsection{Pretraining in Fully Interactive Scenes}
\label{a_ss_pre_exp}

In this section we show the details of the network used in pretraining and the details for policy training in downstream tasks. The network used in pretraining is a UNet structure. The input to the UNet is an RGB image of size $128\times 128$, and the output is a binary mask indicating which area is interactable. The UNet consists of an encoder with ResNet9 architecture, 4 (Upsampling, Conv) blocks and finally a readout module consisting of 2 Conv layers predicting the binary mask.

\begin{figure}
    \centering
    \begin{subfigure}[b]{\columnwidth}
    \includegraphics[width=0.98\columnwidth]{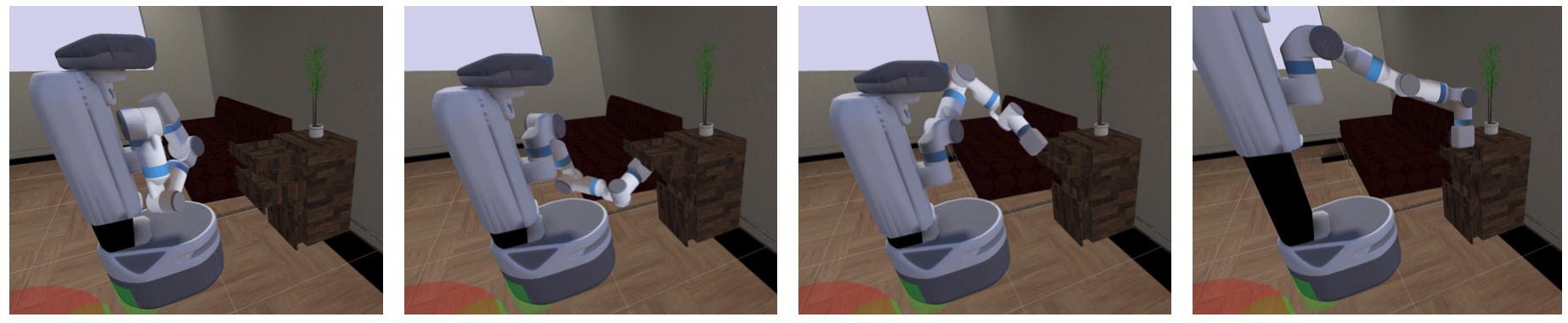}
    \caption{\texttt{PushDrawer} task}
    \end{subfigure}
    
    \begin{subfigure}[b]{\columnwidth}
    \includegraphics[width=0.98\columnwidth]{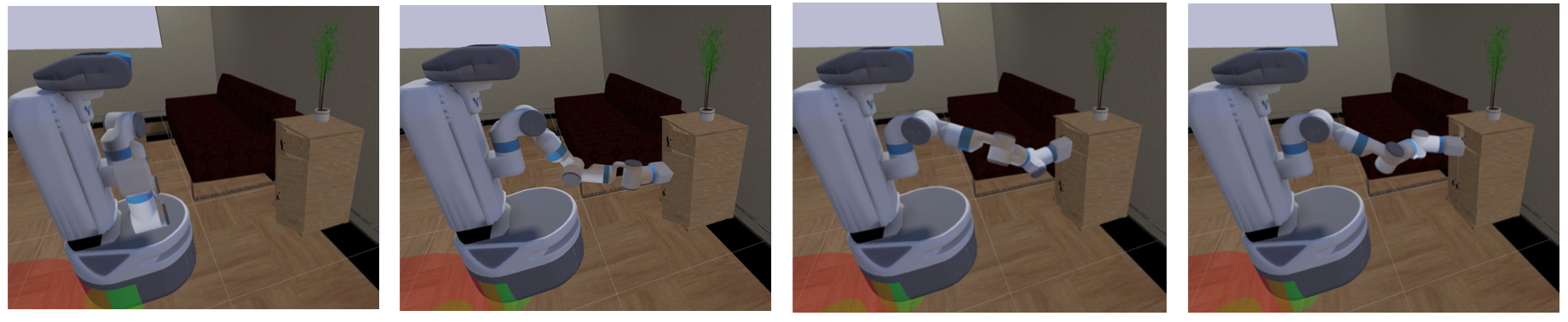}
    \caption{ \texttt{PushCabinet} task} 
    \end{subfigure}
    
    \caption{Execution of (a) \texttt{PushDrawer} task. The agent successfully push the top drawers in. (b) \texttt{PushCabinet} task. The agent push and close the cabinet doors.}
    \label{fig:push_drawer}
\end{figure}

For the downstream task, we focused on pushing task. The observation space is RGB images of size $128\times 128$, and the action space is a point on the image. To give readers an intuitive sense, an example trajectory of the two tasks are shown in Fig.~\ref{fig:push_drawer}. The baseline model for these tasks are DQN with dense Q-value prediction. The policy network uses a 6-layer convolutional neural network to predict an array of Q-values, with the same shape as the input image. The agent picks the pixel with the highest associated Q-value, and use a motion planner to plan a push motion. The method that integrates pretraining modifies the baseline method, by adding a mask to the predicted Q-values with predicted interaction mask, only pixels that are predicted to be interactable keep the Q-values, and the rest are zeroed out. Both setups uses Q-learning algorithm with $\epsilon$-greedy to update the network, we use a learning rate of $1\times 10^{-3}$, discount factor of $0.99$, $\epsilon=0.2$ for exploration, and the policies are trained for $5\times 10^4$ steps.

As shown in the main paper, with pretraining, the agent learns faster.  Fig.~\ref{fig:push_drawer} depicts different stages of the interactions learned by an agent using our learned intermediate representation.

\subsection{Physics Based Rendering}
In this section we include additional information about the shading model and rendering process we use in iGibson. 

{\bf Shading Models} To improve the realism when rendering images of objects, we represent the properties of the surface material of each object using four layers, 1) metallic layer (single channel image), 2) roughness layer (single channel image), 3) albedo (three channel image), and 4) tangent-space normal mapping (three channel image). The information of the layers is combined by our physics-based rendering model. To generate correct light effects on the objects' surfaces, we create environment maps that light the scene. We also pre-integrate the Cook-Torrance specular bidirectional scattering distribution function (BRDF) for varying roughness and viewing directions to accelerate and enable real time rendering. The results are saved into 2-dimensional look up table texture, which is used to scale and add bias to Fresnel reflectance at normal incidence($F_0$) during rendering. The $F_0$ is then used to calculate Shlick's approximation of the Fresnel factor as follows:
$$
F(\theta) = F_0 + (1-F_0)(1-\cos{\theta}) ^ 5
$$

To calculate the specular highlights, we pre-filter environment cube map using GGX normal distribution function importance sampling. The results are saved as an image for later faster retrieval. For diffuse effect we use quasi Monte Carlo sampling with Hammersley sequence to approximate the integration.

{\bf Light Probe Generation} To generate light probes, we use \href{https://www.blender.org/}{Blender} to bake high resolution high dynamic range(HDR) environment textures within iGibson scenes. The light sources are artistically designed and placed on the ceiling. 

{\bf Shadow Mapping} For generating shadows we use shadow mapping techniques and simulate orthogonal uniform light in $+z$ direction. All the objects in the scenes are shadow casters except for the ceiling. The shadows don't necessarily match the real world indoor lights, but create a realistic enough effect that help perceiving depth. 

\subsection{iGibson Performance}

In this section we evaluate the speed of the iGibson simulation environment, analyzing the time for rendering and physics simulation, and its combination.

\subsubsection{Rendering Performance}

We evaluate the performance and speed of our novel iGibson renderer with different working settings. There are many configurable options in iGibson to control the rendering quality. We benchmark two typical use cases. The first one is Reinforcement Learning (\texttt{VisualRL}), where the goal is to generate simulated sensor data (low resolution) as fast as possible. The second one is when higher quality images, as photorealistic as possible, are required, but maintaining a minimum level of rendering speed (\texttt{HighFidelity}). This use case is common when training perceptual solutions (e.g., segmentation, object detection) and also to support collecting human demonstrations with photorealistic enough images and effects such as shadows.

For the first use case, \texttt{VisualRL}, we render $128\times 128$ images, with physically based rendering on, multi sample anti-aliasing off, and shadow mapping off. For the second case, \texttt{HighFidelity}, we render $512\times 512$ images, with physically based rendering on, multi sample anti-aliasing on, shadow mapping on. The results of our benchmark of the rendering time for different modalities with these settings are shown in Table~\ref{tbl:rendering_speed}. 

\begin{table}[t]
  \begin{center}
    \caption{Rendering Speed of iGibson Renderer [fps]   \label{tbl:rendering_speed}
    }
  \begin{tabular}{lccccc} 
  \toprule
  Preset & Modality  & Mean & Max & Min \\
  \midrule
    \multirowcell{6}{\texttt{VisualRL}} & RGB & 409.8 & 1142.7 & 270.4 \\
     & Normal & 530.9 & 1142.0 & 283.7\\
     & Point Cloud & 530.3 & 1129.0 & 282.1\\
     & Semantic Mask & 529.4 & 1140.4 & 281.7\\
     & Optical Flow & 528.1 & 1129.1 & 281.6\\
     & Scene Flow & 526.4 & 1129.9 & 280.1 \\
  \midrule
   \multirowcell{6}{\texttt{HighFidelity}}  & RGB & 188.4 & 289.3 & 114.3\\
     & Normal & 219.6 & 310.1 & 148.9 \\
     & Point Cloud & 240.4 & 345.1 & 174.7\\
     & Semantic Mask & 221.5 & 313.8 & 163.5\\
     & Optical Flow & 240.1 & 348.9 & 170.8\\
     & Scene Flow &  240.5 & 345.9 & 173.9\\
   \bottomrule
  \end{tabular}
  \end{center}
\end{table}

\begin{table}[t]

  \begin{center}
    \caption{Simulator Step Speed and Full Step Speed [\SI{}{Hz}] \label{tbl:phys_speed}}
  \centering
  \begin{tabular}{lccccc} 
  \toprule
  & Robot &  Mean & Max & Min \\
  \midrule
 \multirow{2}{*}{Simulator Physics Step}  & \multirow{1}{*}{With robot} & 175 & 311 & 130 \\
   & \multirow{1}{*}{Scene only}  & 310 & 797 & 92 \\
  \midrule
   \multirow{2}{*}{Simulator Full Step} & \multirow{1}{*}{With robot} & 100 & 150 & 68 \\
   & \multirow{1}{*}{Scene only}  & 136 & 230 & 84 \\
   \bottomrule
  \end{tabular}
  \end{center}
\end{table}

\subsubsection{Physics Simulator and Full Simulator Performance}

The size and number of objects included in our scenes are beyond what is typically included in other simulator or other projects based on PyBullet. We benchmarked the performance of the physics simulator and the final performance of iGibson, combining physics and rendering. In our experiments, the physics simulation timestep is set to \SI{1/120}{\second}. Each simulator step includes four steps of the physics simulator and one rendering pass, corresponding to rendering at 30 fps in simulation. The results of our analysis are shown in Table.~\ref{tbl:phys_speed}. We achieve an average of \SI{100}{Hz} with robot for simulator full step, and \SI{136}{Hz} without robot, which correspond to $3.33\times$ and $4.53\times$ real-time respectively.
 
\subsection{Integration of Additional Datasets of Scenes}
\label{a_aaa}

\subsubsection{Integration of CubiCasa5K}
\label{a_cc}
\begin{figure}
    \centering
    \includegraphics[width=0.98\columnwidth]{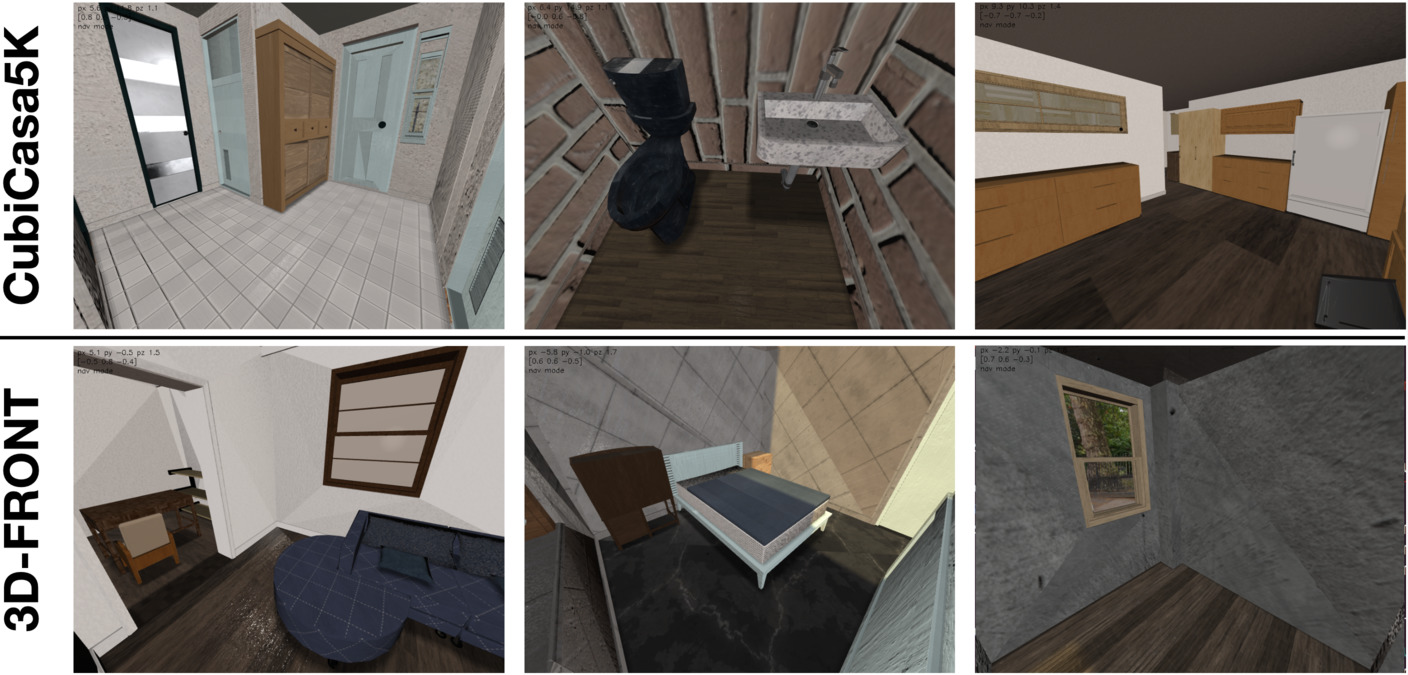}
    \caption{Examples of iGibson scenes created based on the CubiCasa5K and 3D-Front annotations. We provide over 12000 additional fully interactive iGibson scenes based the two datasets.}
    \label{fig:cubicas}
\end{figure}

CubiCasa5K~\cite{kalervo2019cubicasa5k} is a dataset of five thousand annotated floor plans of real world homes in Finland. The annotated floor plans are semi-automatically generated and include the structural elements (walls, doors, windows) and position and size of fixed furniture items (closets, toilets, benches, embedded cabinets, counters, sinks, \ldots). We convert this annotations into iGibson 3D fully interactive scenes with a two step procedure: 1) generate a building based on the annotation of structural elements, and 2) populate the building with object models from our dataset based on the description of poses and sizes from CubiCasa5K. Some of the floor plans in CubiCasa5K included two separate floors; since we do not include outdoor navigation in iGibson, we split them into two separate indoor scenes. In total, we offer 6297 scene in iGibson based on the real-world layouts of CubiCasa5K. Fig.~\ref{fig:cubicas} depicts some examples of the scenes included created based on the CubiCasa5K dataset.

\subsubsection{Integration of 3D-Front}
\label{a_tdf}

3D-Front~\cite{fu20203d} is a large dataset of layouts with room models populated with furniture. The layouts have been created by artists and interior designers. It includes 18,797 rooms, around 10,000 houses, and 7,302 furniture models. We convert 3D-Front static scenes into iGibson 3D fully interactive scenes with a two step procedure: 1) we keep the original building structural meshes as visual meshes while procedurally generating collision meshes for structural elements that approximate their geometry, and 2) we populate the building with object models from our dataset based on the description of poses and sizes from 3D-Front.

There are four challenges we faced when integrating 3D-Front to convert their scenes into interactive scenes for iGibson. First, some 3D-Front scenes contain objects of undefined categories, corresponding to one-of-a-kind pieces. We skip these scenes, since we cannot generate appropriate low-poly collision meshes for these objects. 
Second, some of the 3D data within 3D-Front is corrupted or contain shape errors (see reported issue \href{https://github.com/3D-FRONT-FUTURE/3D-FRONT-ToolBox/issues/2#issuecomment-682678930}{in this link}). 
Third, the kitchen cabinets in 3D-Front are not annotated as objects, but instead the entire kitchen furniture is a single object with each panel of the furniture (front and lateral panels, internal shelves) annotated as a separate part. This impedes us to generate interactive versions of the kitchen cabinets. We include two alternative versions of the scenes: a) a version with non-interactive kitchen cabinets, and b) a version without any kitchen cabinets. We expect this problem to be solved in future annotations of 3D-Front. Fourth, while 3D-Front dataset includes a layout description of rooms and elements, including their position and size, the furniture pieces are sometimes defined as overlapping significantly with each other (see reported issue \href{https://github.com/3D-FRONT-FUTURE/3D-FRONT-ToolBox/issues/4}{in this link}). This has a severe effect in our physics simulation as it tries to solve the penetrating contact. To alleviate this issue, we remove objects that overlap more than 80\% of volume with others. For objects that overlap less than 80\%, we reduce their size (with a threshold of reducing no more than 20\%) until the overlapping is resolve. After all the aforementioned filtering and fixing processes, we obtained 6049 scenes with correct object placements without overlapping, ready to be used in interactive tasks.

\subsubsection{Comparing iGibson, CubiCasa5K and 3D-Front Scenes}

There are two significant differences between the 15 fully interactive iGibson scenes, and the ones obtained from the integration of Cubicasa5K and 3D-Front. The first one is in density of objects. Our iGibson curated scenes contain a density of 75 objects per room. This is more realistic and significantly higher density that the one in 3D-Front rooms (37) and CubiCasa5K rooms (32). A second difference between the scenes is in the albedo and material effects of walls and floors. In the 15 iGibson scenes provided, we bake the effects of the ambient lighting and the light sources on walls, floors and ceilings into a RGB texture map. This provides a more realistic lighting effect. Differently, the structural elements of CubiCasa5K and 3D-Front do not present this enhanced diffuse color channel. To bake lighting sources in CubiCasa5K and 3D-Front scenes, we need lighting information that are not contained as part of the datasets. It would require a manual annotation of light location and strength in thousands of scenes, an effort that is beyond our attempt to make these datasets available to the community. Despite these differences, we believe getting access to these two large datasets of scenes significantly complement the lower number (15) but higher quality scenes provided with iGibson.

\balance

\end{document}